\documentclass[conference]{IEEEtran}

\usepackage{amsmath,amssymb,amsfonts}
\usepackage{graphicx}
\usepackage{booktabs}
\usepackage{multirow}
\usepackage{subcaption}
\usepackage{url}

\usepackage[breaklinks,colorlinks,citecolor=blue,urlcolor=blue,linkcolor=blue]{hyperref}
\usepackage{flushend}  

\usepackage[capitalize,noabbrev]{cleveref}

\usepackage{xspace}
\makeatletter
\DeclareRobustCommand\onedot{\futurelet\@let@token\@onedot}
\def\@onedot{\ifx\@let@token.\else.\null\fi\xspace}
\makeatother

\def\eg{\emph{e.g}\onedot}

\def\ie{\emph{i.e}\onedot}

\def\etal{\emph{et al}\onedot}

\graphicspath{{./}}

\begin{document}

\title{A Camera-Native Talking-Head Video Dataset for Various Computer Vision Tasks}

\author{
\IEEEauthorblockN{Babak Naderi, Ross Cutler, Nabakumar Singh Khongbantabam}
\IEEEauthorblockA{Microsoft Corporation, Redmond, USA}
}

\maketitle

\begin{abstract}
Talking-head videos constitute a predominant content type in real-time
communication, yet publicly available datasets for video processing
research in this domain remain scarce and limited in signal fidelity.
In this paper, we open-source a camera-native dataset of 838 talking-head
recordings
(approximately 210 minutes), each 15\,s in duration, captured from 799
participants across 443 camera-code categories in their natural
environments.
All recordings are stored using the FFV1 lossless codec, preserving the
camera-native signal---uncompressed (24.7\%) or MJPEG-encoded
(75.3\%)---without additional lossy processing.
Each recording is annotated with a Mean Opinion Score (MOS) and ten perceptual
quality tokens that jointly explain 64.4\% of the MOS variance.
From this corpus, we curate a stratified benchmarking subset of
120 clips in three content conditions: original, background blur, and
background replacement.
Codec efficiency evaluation across four datasets and four codecs, namely H.264, H.265, H.266, and AV1, yields
VMAF BD-rate savings up to $-71.3\%$ (H.266) relative to H.264, with
significant encoder$\times$dataset ($\eta_p^2 = .112$) and
encoder$\times$content condition ($\eta_p^2 = .149$) interactions,
demonstrating that both content type and background processing affect
compression efficiency.
A preliminary super-resolution evaluation with four SR models confirms that
the dataset significantly affects absolute performance while preserving
model rankings, demonstrating applicability beyond codec benchmarking.
The dataset offers 5$\times$ the scale of the largest prior talking-head
webcam dataset (838 vs.\ 160 clips) and preserves the camera-native signal
without additional lossy compression,
establishing a resource for benchmarking video
compression, super-resolution, quality assessment, and enhancement
models in real-time communication.
\end{abstract}

\begin{IEEEkeywords}
Talking-head video, Video quality assessment, Lossless video dataset, Codec evaluation
\end{IEEEkeywords}

\section{Introduction}
\label{sec:intro}

Video conferencing has become a primary mode of remote collaboration,
with talking-head video constituting a predominant content type in
real-time communication (RTC) platforms.
The perceptual quality of this video directly affects communication
effectiveness: degraded video reduces social presence, hampers non-verbal
cue interpretation, and diminishes perceived meeting
quality~\cite{skowronek_qoe_2022}.
Improving the video processing pipeline---through better compression,
restoration, or enhancement---therefore has a direct impact on the
user experience of RTC participants.

Despite this importance, research on video processing tasks for
talking-head content, including lossy compression, super-resolution (SR),
and denoising, relies on datasets that either lack domain specificity or
introduce compression artifacts during
capture~\cite{naderi_vcd_2024,xie2022vfhq}.
Reliable evaluation of these tasks requires camera feeds that preserve
authentic scene-level degradations (noise, low-light conditions, motion
blur) while avoiding post-processing artifacts such as lossy compression,
spatial rescaling, or frame interpolation that would confound the
measurement of algorithm performance.
To our knowledge, no public dataset combines domain-representative webcam
capture, lossless encoding, and the scale required for systematic benchmarking
and model development in the RTC domain.
We substantiate this gap below by reviewing the relevant dataset landscape.

Talking-head content exhibits distinct spatial and temporal properties
that differentiate it from general video~\cite{naderi_vcd_2024}.
Backgrounds are typically static with low spatial complexity, faces
require accurate texture preservation for fine detail, and subtle temporal
variations from speech and gestures are sensitive to compression-induced
artifacts.
Standard codec test sequences~\cite{bossen_common_2013,bossen_jvet_2019}
and widely used benchmarks such as
UVG~\cite{mercat_uvg_2020} and MCL-JCV~\cite{wang_mcl-jcv_2016}
contain professionally captured content that does not reflect these
properties.

Among talking-head corpora, large-scale web-sourced datasets provide
scale and identity diversity but limited signal fidelity.
VoxCeleb~\cite{nagrani_voxceleb:_2017} and
VoxCeleb2~\cite{chung_voxceleb2:_2018} together aggregate over one million
utterances from YouTube, but their quality is bounded by platform
compression~\cite{xie2022vfhq}.
HDTF~\cite{zhang2021hdtf} and CelebV-HQ~\cite{zhu2022celebvhq} curate
higher-resolution face tracks from online video, yet remain subject to
platform encoding.
DH-FaceVid-1K~\cite{di2024dhfacevid1k} provides approximately 270{,}000
clips with multi-modal annotations but standardizes all content to
$512\times512$ face crops, discarding background context relevant to video
conferencing.

Controlled and specialized datasets address different limitations.
MEAD~\cite{wang2020mead} offers controlled multi-view studio capture at the
cost of ecological validity for RTC scenarios, while
FaceForensics++~\cite{rossler2019faceforensics} targets manipulation
detection rather than providing pristine references.
VFHQ~\cite{xie2022vfhq}, among the most relevant datasets for video
face SR, curates over 16{,}000 high-fidelity clips and demonstrates that
models trained on web-compressed video reproduce compression artifacts
rather than recover genuine detail; however, VFHQ content is still
derived from web video and does not provide lossless capture.
Naderi~\etal~\cite{naderi_vcd_2024} introduced the first public dataset
targeting video conferencing for codec evaluation, comprising 160 clips of
10\,s duration with desktop and mobile recording scenarios and background
processing variants.
However, those recordings were captured using the lossy output of consumer
webcams (\eg, VP8 or H.264), embedding compression artifacts into the
reference signal.

Beyond compression evaluation, these limitations also affect
super-resolution research, where reference quality directly determines
model fidelity.
General-purpose SR benchmarks such as REDS~\cite{nah_ntire_2019} provide
standardized evaluation tracks, and models such as
EDVR~\cite{wang2019edvr} and
BasicVSR/BasicVSR++~\cite{chan_basicvsr_2021,chan_basicvsr_2022} serve as
widely adopted baselines.
Real-world SR datasets address the realism gap:
RealVSR~\cite{yang2021realvsr} captures paired sequences with two cameras
at different focal lengths, and
VideoLQ~\cite{chan_investigating_2022} provides a benchmark for blind
real-world video SR.
However, these datasets focus on general scenes and do not capture the
domain-specific characteristics of webcam talking-head video, such as
webcam sensor noise profiles, auto-exposure behavior, and RTC-typical
background processing.
Chan~\etal~\cite{chan_investigating_2022} corroborate that models trained
on pre-compressed video reproduce compression artifacts rather than
recover genuine detail, suggesting that pre-compressed references may
similarly confound codec evaluation.
Emerging compression paradigms further motivate lossless talking-head data:
Generative Face Video Coding (GFVC)~\cite{chen2024gfvc_survey} benefits
from clean identity and lip-motion ground truth, and neural RTC compression
systems such as Gemino~\cite{sivaraman2024gemino} combine SR with
compression, making uncompressed references essential for isolating codec
artifacts from camera noise.

We present a dataset of 838 talking-head recordings, each 15\,s
in duration (approximately 3.5 hours total), captured from 799 participants
across 443 camera-code categories in their natural
environments.
The capture application opens each camera at its largest supported resolution
and selects the highest-quality pixel format available.
The priority order is YUYV 4:2:2 or NV12 4:2:0 (uncompressed), with MJPEG
as a fallback.
All frames are encoded with the FFV1 lossless codec.
When the camera exposes only lossy formats (\eg, MJPEG), the compressed
stream is decoded and losslessly stored, preserving camera-compressed quality
without further degradation from the capture pipeline.
We refer to this capture approach as ``camera-native,'' as the capture
pipeline applies no lossy compression beyond camera firmware processing
(details in \cref{sec:data_collection}).

Each recording is annotated with a Mean Opinion Score (MOS) obtained via
subjective testing using the Absolute Category Rating (ACR)
method according to the ITU-T Rec. P.910~\cite{itu-t_recommendation_p910_subjective_2023}
and with ten perceptual quality attributes (\eg, blur, noise, low resolution,
lighting issues) derived through a three-phase crowdsourced annotation process.
From the full corpus, we curate a benchmarking subset of 120 clips stratified
by Spatial Information (SI), Temporal Information (TI), and MOS.
The subset is organized into three groups, each containing unique clips:
original talking-head clips (TH), clips with commercial-grade background
blur (TH-BB), and clips with background replacement (TH-BR).

We summarize our contributions as follows:
\begin{enumerate}
    \item A diverse, camera-native talking-head webcam video dataset
    comprising 838 losslessly encoded recordings from 799 participants across
    443 camera-code categories, with four recording scenarios covering
    common video conferencing behaviors.
    \item A multi-dimensional quality annotation scheme combining ACR-based MOS
    with ten perceptual quality tokens, validated through cross-study reliability
    analysis (Pearson $r \geq 0.859$ between independent annotation studies).
    \item A stratified benchmarking subset of 120 clips in three groups (TH,
    TH-BB, TH-BR), balanced across quality levels, spatial--temporal complexity,
    and distortion types.
    \item An evaluation of the dataset's utility for codec compression
    efficiency analysis across H.264 (AVC)~\cite{wiegand_overview_2003},
    H.265 (HEVC)~\cite{sullivan_overview_2012},
    H.266 (VVC)~\cite{bross_overview_2021}, and
    AV1~\cite{han_technical_2021},
    including a VP8 contamination experiment quantifying the effect of
    platform-level capture compression on codec benchmarking,
    and a preliminary super-resolution evaluation demonstrating the
    dataset's applicability for SR model benchmarking
    (see \cref{sec:analysis}).
\end{enumerate}


\section{Dataset}
\label{sec:dataset}

This section describes the data collection methodology, the composition of
the published dataset, the quality annotation process, and the construction
of the benchmarking subset.

\subsection{Data Collection}
\label{sec:data_collection}

We developed a custom recording application built on the
DirectShow\footnote{\url{https://learn.microsoft.com/en-us/windows/win32/directshow/}}
multimedia framework and FFmpeg\footnote{\url{https://ffmpeg.org/}}.
The application interfaces with webcams via the USB Video Class (UVC)
protocol and captures video directly from the camera hardware with minimal
software processing.
Participants installed the application on their personal computers and
recorded themselves in their natural environments (\eg, home offices, living
rooms), ensuring authentic and diverse capture conditions.

The application opens each camera at its largest supported resolution
(minimum 1280$\times$720 at 30\,fps, 16:9 aspect ratio) and selects the
highest-quality pixel format available: uncompressed YUYV422 or NV12 when
supported, with Motion JPEG (MJPEG) as a fallback.
In the published dataset, 24.7\% of recordings use uncompressed formats
(YUYV422 or NV12) and 75.3\% use MJPEG, reflecting the limited support
for uncompressed output at high resolutions in consumer webcam hardware.
All frames are encoded with the FFV1 lossless video
codec~\cite{niedermayer_ffv1_2022}, which guarantees bit-exact
reconstruction, and frame timestamps are passed through from the
camera without interpolation, dropping, or duplication.

Each recording session captures a 20-second window, of which the first
5\,s serve as pre-roll (discarded) and the remaining 15\,s constitute the
published recording.
Participants review the playback and accept or reject the recording;
rejected recordings are deleted and retaken.
We refer to this capture approach as ``camera-native.''
Camera firmware applies demosaicing, white balance, and gamma correction
before the signal reaches the recording software, but no further lossy
compression is applied by the capture pipeline.
This design eliminates the double-compression artifacts
(\ie, camera compression followed by capture-software compression)
present in conventional webcam recording
workflows~\cite{naderi_vcd_2024}.
The published videos, benchmarking subset, and associated metadata are
open-sourced at \url{https://github.com/microsoft/VCD}.
Additional technical details on pixel format handling, lossless encoding
configuration, and frame buffering are provided in the supplementary
material.

\subsection{Dataset Composition}
\label{sec:composition}

Participants were recruited via the Prolific crowdsourcing
platform and recorded themselves performing one of four randomly assigned
scenarios designed to elicit diverse motion patterns:
continuous slow body movement (S01, 314 clips), hand counting exercise
(S02, 154 clips), text reading exercise (S03, 300 clips), and natural
video call behavior (S04, 70 clips).
S01 and S03 received higher assignment probability to provide larger
sample sizes for the two scenarios with the most controlled motion
characteristics.
These scenarios introduce diverse temporal characteristics, from
relatively low motion in S03 to continuous upper-body motion in S01.

A total of 1{,}119 recordings were collected.
After quality control and final publication filtering, 838 clips from 799
unique participants are published.
Quality-control exclusions include low frame rate (below 15\,fps), frame
drops, perceptible blockiness detected through annotation, and other
technical failures.
The dataset spans 443 camera-code categories: 442 USB vendor/product
identifier combinations and one category for recordings without an available
camera code.
These categories include integrated laptop cameras and external USB webcams
from multiple manufacturers.
\Cref{tab:resolution_format} shows the distribution of recordings across
resolutions and pixel formats.
The majority of clips are captured at 720p or 1080p, reflecting the
capabilities of participants' consumer webcams, with higher-resolution
captures (1440p, 4K) representing a small fraction.
The participant pool comprises 63\% male and 37\% female contributors;
full dataset statistics including demographics are provided in the
supplementary material.

\begin{table}[tb]
\centering
\caption{Distribution of recordings by resolution and input pixel format
(\% of 838 published clips).
Uncompressed combines YUYV422 and NV12 formats.
Percentages are independently rounded and may not sum exactly.}
\label{tab:resolution_format}
\begin{tabular}{@{}lccc@{}}
\toprule
Resolution & Uncompressed & MJPEG & Total \\
\midrule
720p  & 10.9\% & 49.9\% & 60.7\% \\
1080p & 11.2\% & 21.8\% & 33.1\% \\
1440p or larger &  2.6\% &  3.6\% &  6.2\% \\
\midrule
Total & 24.7\% & 75.3\% & 100\% \\
\bottomrule
\end{tabular}
\end{table}

\subsection{Quality Annotation}
\label{sec:annotation}

Each recording is annotated along two complementary dimensions using
subjective assessment: an overall quality rating, represented as a Mean
Opinion Score (MOS), and a set of multi-label perceptual quality attributes
(problem tokens).
The overall perceived quality was assessed with the ACR methodology
following ITU-T Recommendation
P.910~\cite{itu-t_recommendation_p910_subjective_2023}, deployed via its
crowdsourcing
implementation~\cite{naderi_crowdsourcing_2024}.
The study was conducted on the Prolific platform with 216 accepted workers,
yielding approximately 7 votes per clip.

Quality control followed established crowdsourcing best
practices~\cite{naderi_crowdsourcing_2024,egger-lampl_crowdsourcing_2014,ribeiro_crowdmos_2011}.
Participants passed qualification checks including Ishihara color vision
plates~\cite{clark_ishihara_1924}, device validation (minimum
1920$\times$1080 display at 30\,Hz), attention checks, trapping questions
with clips of known expected quality, and gold standard clips with
unambiguous quality levels.
Workers failing these checks were excluded from the aggregated ratings.

To provide fine-grained, multi-label diagnostic annotations, we developed
a set of ten perceptual quality tokens through a three-phase process.
In the first phase, 273 crowdsourced assessors described observed
distortions in free-text comments for approximately 260 clips
(a 31\% sample of the dataset), yielding 2{,}596 comments that were analyzed
using keyword-based tagging and large language model (LLM)-assisted
classification (the LLM step was assistive only; the final taxonomy was
human-validated).
The analysis identified blur (42.9\% of comments), lighting/color issues
(23.9\%), and noise (21.0\%) as the most frequently mentioned distortions.
In the second phase, the derived token set was validated on a 200-clip
random subsample with 24 workers providing 5--6 votes per clip; distortion
tokens were presented in randomized order to mitigate primacy and recency
biases.
In the third phase, all 1{,}119 recorded clips (including those later
excluded by quality control) plus 160 reference clips
from prior work~\cite{naderi_vcd_2024} were annotated by 112 workers with
7--8 votes per clip.
A token is considered selected for a clip only if $\geq$2 assessors
independently chose it, reducing noise from idiosyncratic selections.
\Cref{tab:tokens} lists the ten tokens and their selection rates.

\begin{table*}[tb]
\centering
\caption{Perceptual quality tokens and their selection rates across the
838 published clips.
A token is considered selected when $\geq$2 assessors independently
chose it.}
\label{tab:tokens}
\small
\begin{tabular}{@{}p{2.0cm}p{10.0cm}r@{}}
\toprule
Token & Description & Sel.\ (\%) \\
\midrule
Noisy          & Random pixel-level luminance/chrominance fluctuations (sensor noise) & 69.8 \\
Low resolution & Insufficient spatial detail; overall poor definition                  & 61.3 \\
Lighting/color & Over-/under-exposure, direct glare, white balance deviation, color cast & 36.6 \\
No issue       & No perceptible quality degradation                                    & 29.2 \\
Blurry           & Loss of edge sharpness; defocus or motion blur                        & 13.6 \\
Choppy motion  & Irregular temporal motion; stuttering or judder                       &  6.0 \\
Framerate      & Perceptibly low or abnormal frame rate                                &  5.9 \\
Banding        & Visible color banding or scan-line artifacts                          &  5.8 \\
Other          & Uncategorized quality issues                                          &  4.4 \\
Blockiness     & Block-based compression artifacts (8$\times$8 DCT)                    &  0.0 \\
\bottomrule
\end{tabular}
\end{table*}

Cross-study Spearman correlations between token selection rates from the
pilot (Phase~2) and the full annotation (Phase~3) show moderate to strong
agreement ($\rho \geq 0.5$) for seven of ten tokens.
The near-zero correlation for blockiness ($\rho = -0.009$) is consistent
with the lossless capture design: the capture pipeline applies no
additional block-based compression, and the MJPEG encoding performed by
camera firmware operates at sufficiently high quality that block boundaries
are not perceptually salient at typical webcam bitrates.
By contrast, 10.6\% of clips from the VCD dataset~\cite{naderi_vcd_2024},
which uses lossy webcam output, exhibited blockiness in the same
annotation study; in the present dataset, blockiness was detected in
only 15 of 1{,}119 recordings (1.3\%), and those clips were excluded
from the published set.
The MOS values obtained under the problem token paradigm correlate strongly
with the independently collected ACR MOS (Pearson $r = 0.859$ on the full
dataset of 1{,}119 clips, $r = 0.893$ on the 200-clip common subset),
confirming that the additional annotation task does not degrade scalar
quality judgments.

A multiple linear regression of MOS on all ten token proportions yields
$R^2 = 0.644$, indicating that the tokens jointly explain 64.4\% of the
variance in overall quality.
Blur and low resolution have the largest negative coefficients
($\beta \approx -1.0$), followed by noise ($\beta = -0.87$) and motion
($\beta = -0.80$).
The tokens represent largely independent perceptual dimensions
(Kaiser--Meyer--Olkin measure $= 0.475$, below the 0.50 threshold
for factor analysis~\cite{kaiser_index_1974}), suggesting that they
capture non-overlapping quality
information rather than reflecting a smaller set of latent factors.

\begin{figure}[tb]
\centering
\includegraphics[width=0.85\linewidth]{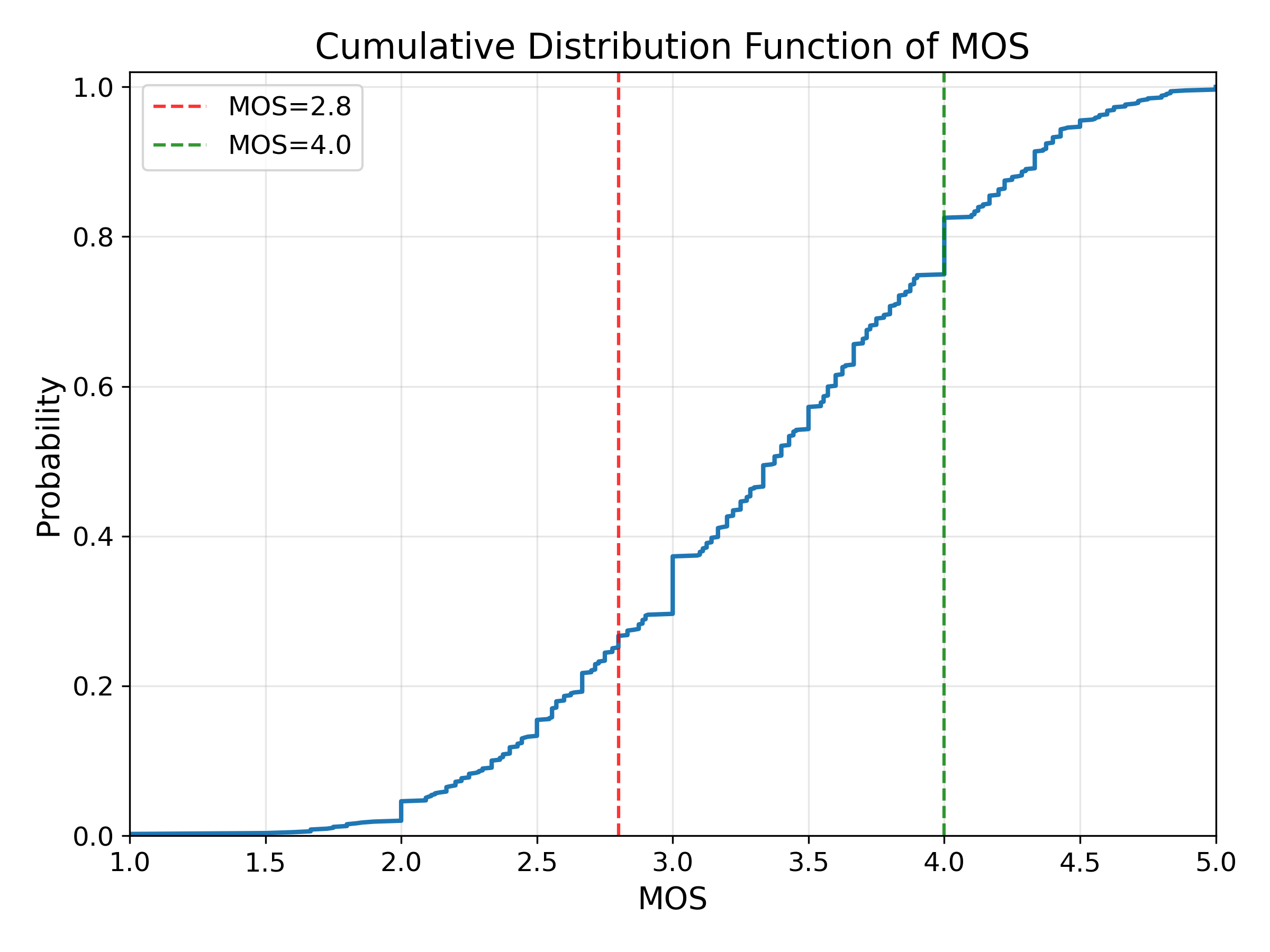}
\caption{Cumulative distribution function of MOS values across the
838 published clips.
}
\label{fig:mos_cdf}
\end{figure}

\Cref{fig:mos_cdf} shows the cumulative distribution function (CDF) of
MOS values across all 838 published clips.
More than 50\% of clips receive a MOS below 3.5, indicating that a
substantial fraction of consumer webcam feeds captured at their maximum
supported resolution contain perceptible quality impairments,
suggesting a practical need for video enhancement in this domain.

\subsection{Benchmarking Subset}
\label{sec:benchmarking}

From the 838 published clips, we curate a benchmarking subset of 120 clips
organized into three mutually exclusive groups of 40 clips each.
Each clip is trimmed to 10\,s to provide a duration suitable for
subjective quality testing.
The three groups serve distinct evaluation purposes:

\begin{itemize}
    \item \textbf{Talking Head (TH):} Original clips without any
    post-processing, representing unmodified webcam-captured content.
    \item \textbf{Talking Head -- Background Blur (TH-BB):} Clips processed
    with a commercial-grade video conferencing background blur pipeline,
    simulating a common real-time processing scenario.
    \item \textbf{Talking Head -- Background Replacement (TH-BR):} Clips
    processed with a commercial-grade video conferencing background replacement
    pipeline using four popular virtual backgrounds, with one background
    randomly assigned per clip.
\end{itemize}

\begin{figure*}[tb]
\centering
\begin{subfigure}[b]{0.32\linewidth}
  \includegraphics[width=\linewidth]{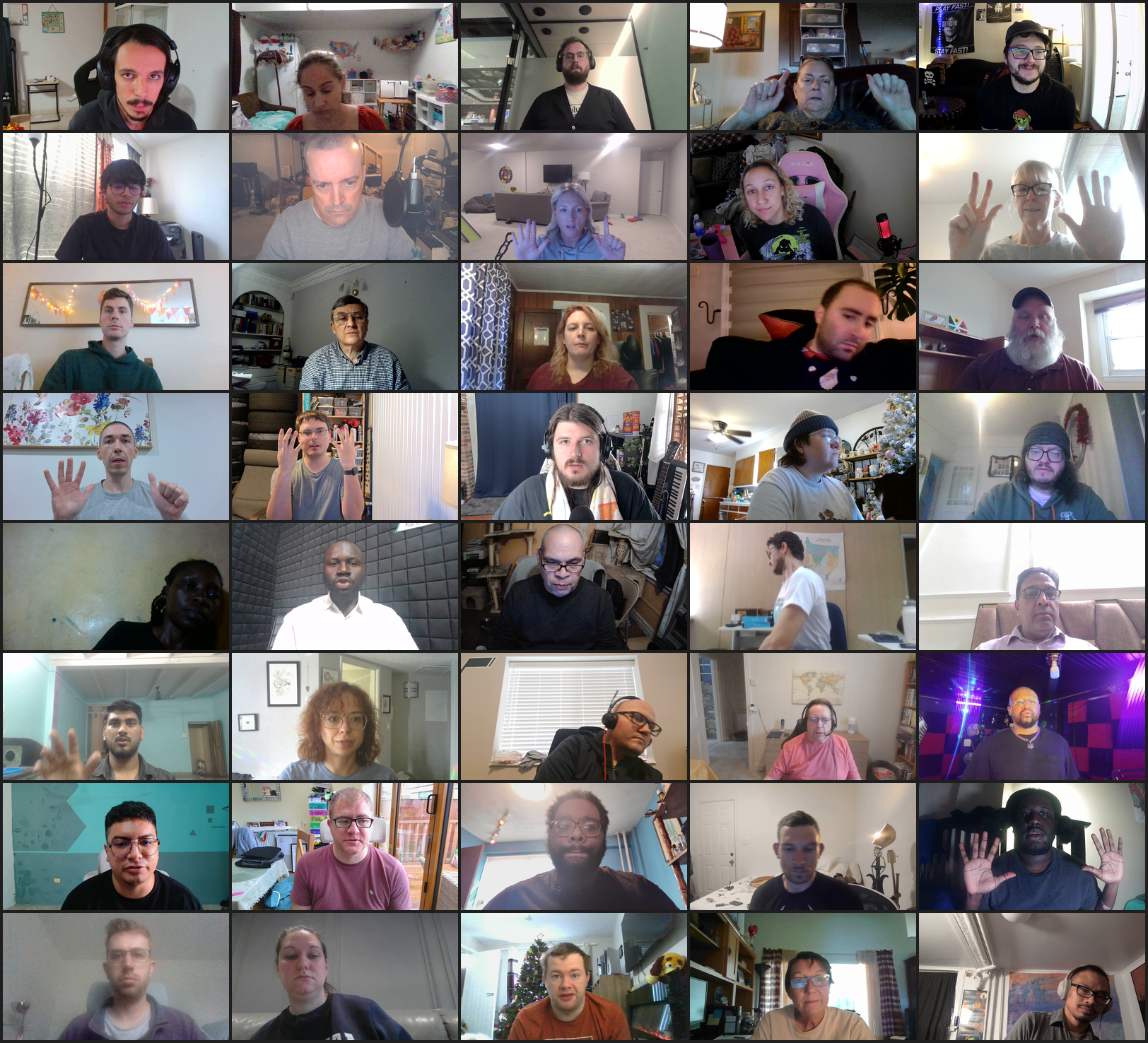}
  \caption{TH}
  \label{fig:atlas_th}
\end{subfigure}
\hfill
\begin{subfigure}[b]{0.32\linewidth}
  \includegraphics[width=\linewidth]{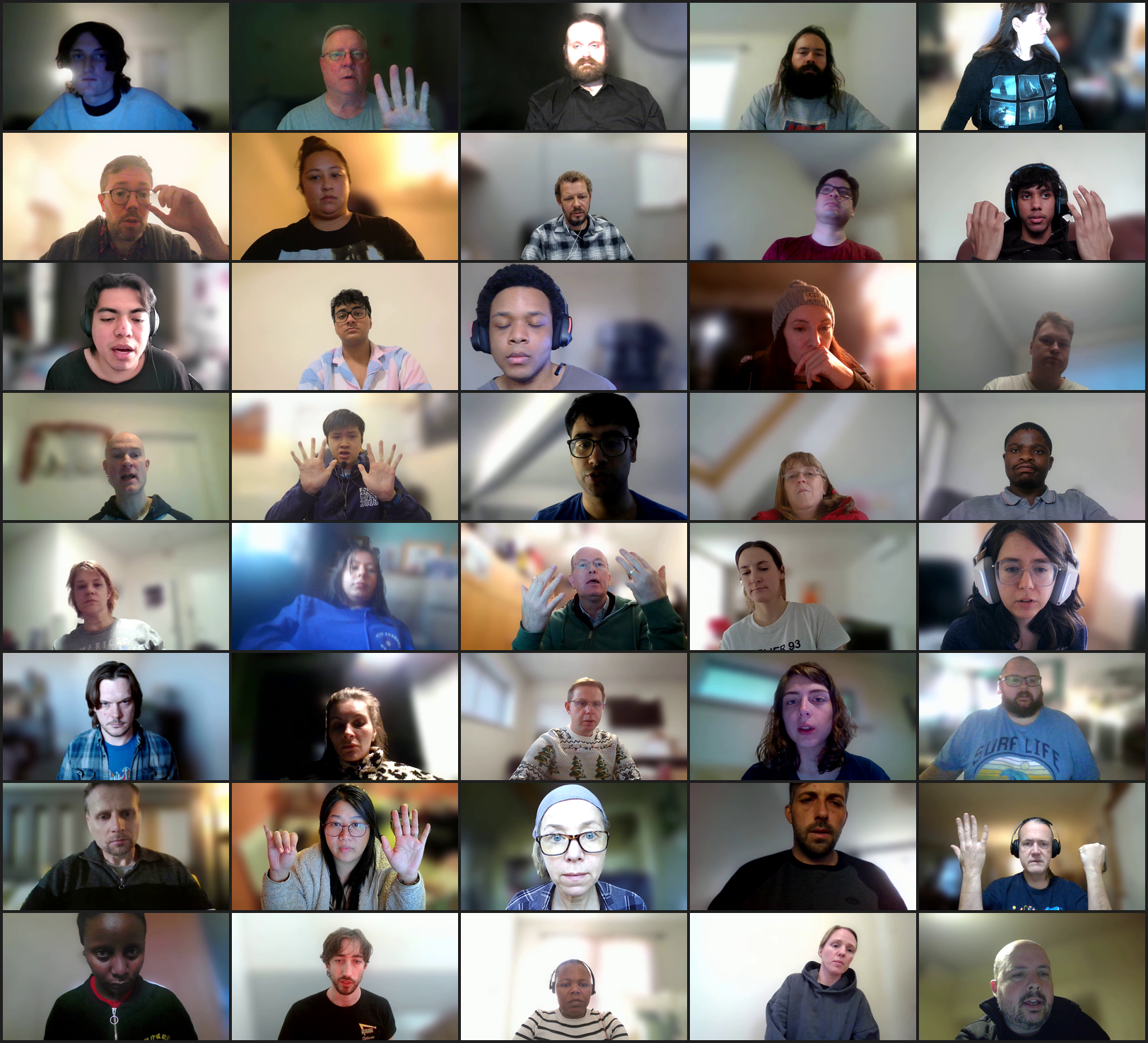}
  \caption{TH-BB}
  \label{fig:atlas_bb}
\end{subfigure}
\hfill
\begin{subfigure}[b]{0.32\linewidth}
  \includegraphics[width=\linewidth]{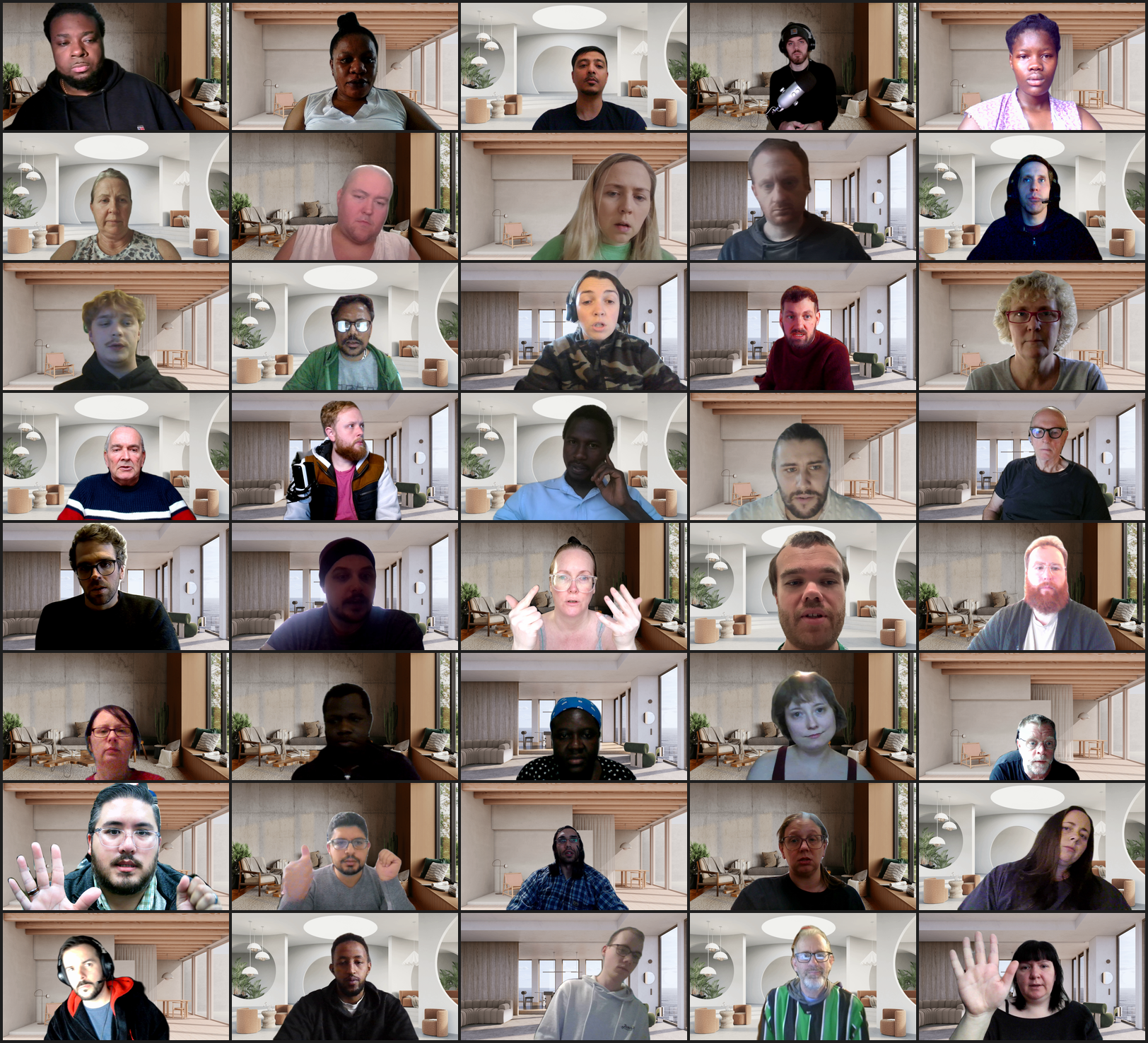}
  \caption{TH-BR}
  \label{fig:atlas_br}
\end{subfigure}
\caption{Thumbnail atlases of the three benchmarking groups. (a) Talking Head, (b) Talking Head - Background Blur, (c) Talking Head - Background Replacement}
\label{fig:atlas}
\end{figure*}

\Cref{fig:atlas} shows thumbnail atlases for clips from
each of the three benchmarking groups.

\subsubsection{Stratification strategy.}
\label{sec:stratification}

The subset selection employs a stratified sampling algorithm to ensure that
each group covers the full range of quality levels, spatial--temporal
complexity, and distortion types.
Clips are assigned to one of 12 strata formed by the Cartesian product of
three MOS bins (Low: $[1.0, 2.8)$, Medium: $[2.8, 4.0)$, High:
$[4.0, 5.0]$) and four SI$\times$TI quadrants (split at the population
medians of SI and TI, computed per ITU-T
P.910~\cite{itu-t_recommendation_p910_subjective_2023}).
The target MOS distribution follows a 25\%/50\%/25\%
(Low/Medium/High) ratio, mirroring the population distribution which
is skewed toward medium quality.

A greedy scoring heuristic selects clips iteratively, balancing five
objectives through a weighted composite score: (1)~stratum quota
fulfillment, (2)~coverage of rare perceptual quality tokens (each of
the 10 tokens should appear $\geq$2 times per group),
(3)~feature-space diversity in normalized MOS/SI/TI coordinates,
(4)~participant uniqueness within and across groups (no participant
appears more than once per group), and (5)~a soft preference for manually curated clips that exhibit sufficient
structural detail to challenge video processing pipelines.
Groups are built sequentially with strict mutual exclusivity.
The resulting 120 clips represent an approximately 14\% sampling rate from
the 838 eligible recordings.

\begin{figure*}[tb]
\centering
\begin{subfigure}[b]{0.32\linewidth}
  \includegraphics[width=\linewidth]{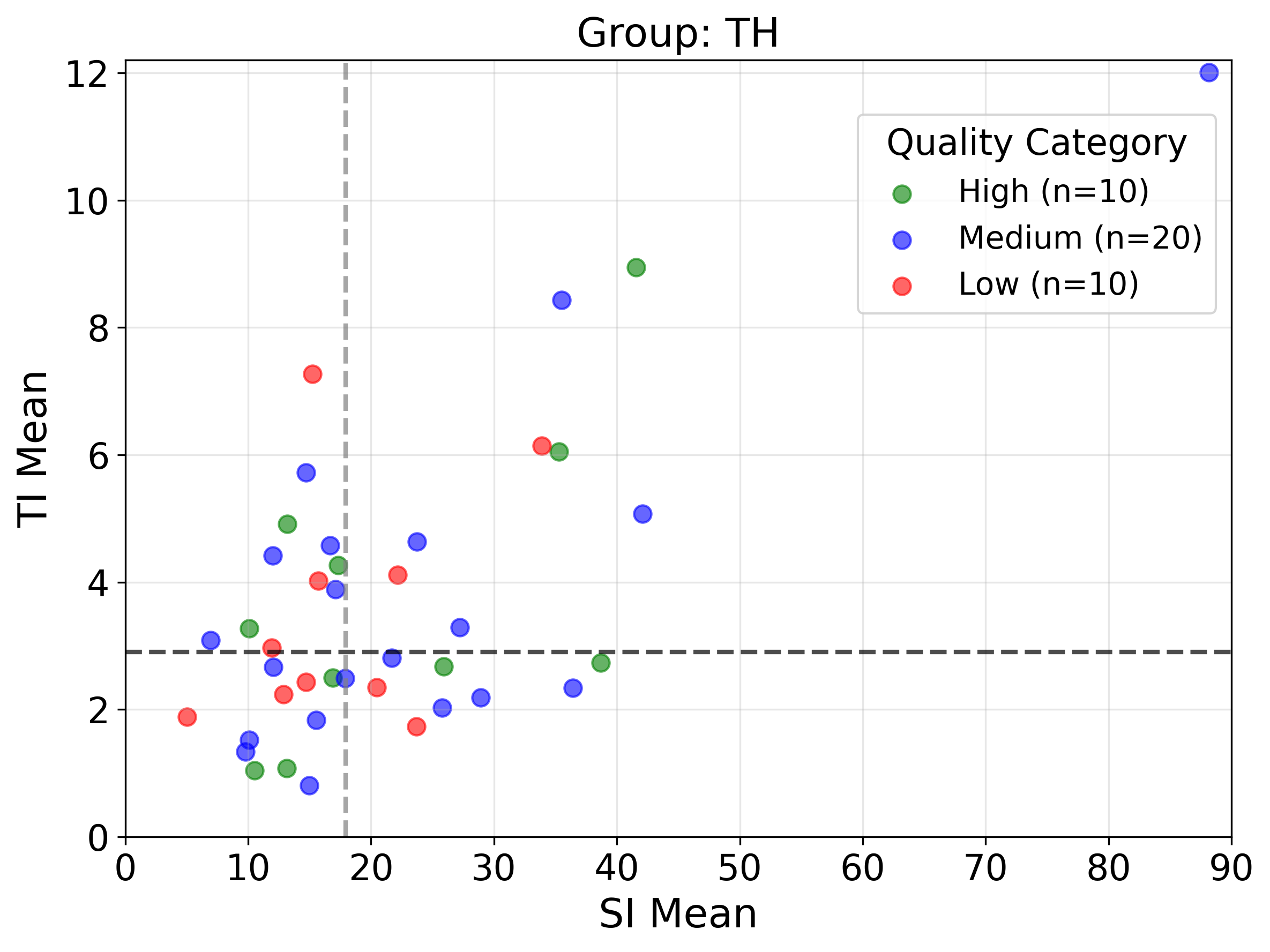}
  \caption{TH}
  \label{fig:siti_th}
\end{subfigure}
\hfill
\begin{subfigure}[b]{0.32\linewidth}
  \includegraphics[width=\linewidth]{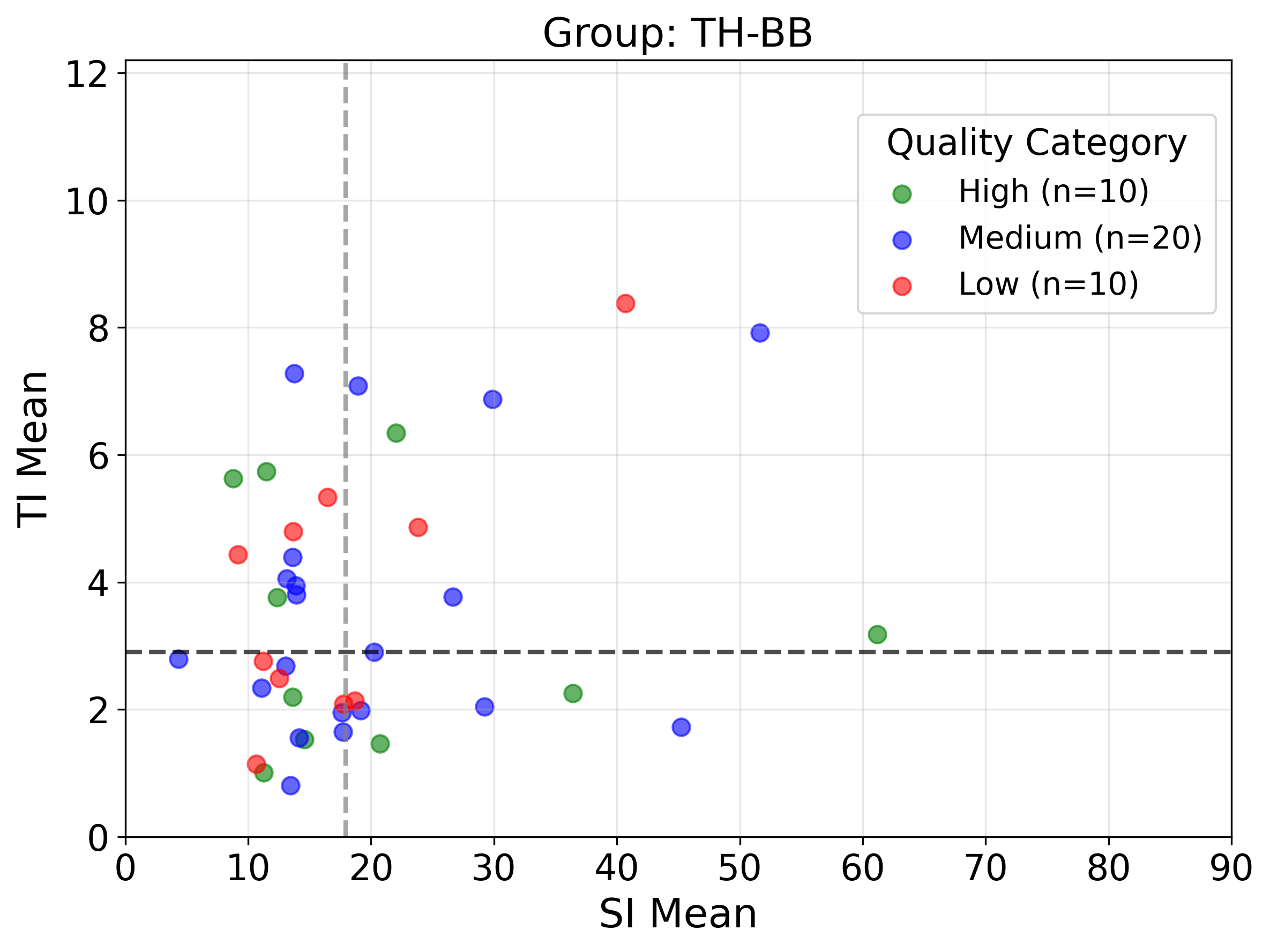}
  \caption{TH-BB}
  \label{fig:siti_bb}
\end{subfigure}
\hfill
\begin{subfigure}[b]{0.32\linewidth}
  \includegraphics[width=\linewidth]{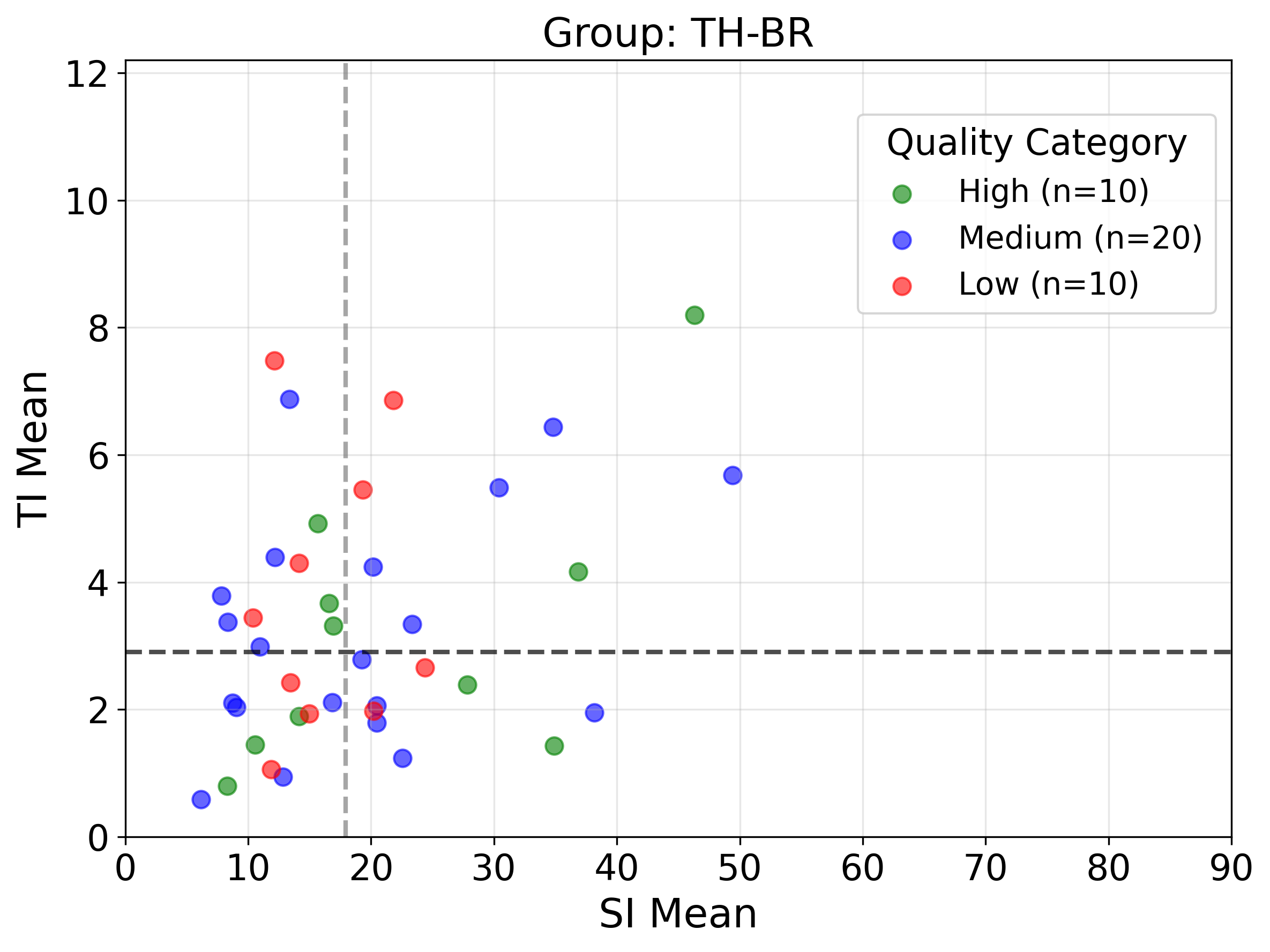}
  \caption{TH-BR}
  \label{fig:siti_br}
\end{subfigure}
\caption{Distribution of clips in the SI--TI space for each benchmarking
group, color-coded by MOS category (green: High, blue: Medium, red: Low).
Dashed lines indicate population medians.}
\label{fig:groups_siti}
\end{figure*}

\Cref{fig:groups_siti} shows the distribution of clips in each group
across the SI--TI space, color-coded by quality category.
The spread across all four quadrants confirms that the stratification
achieves the intended coverage of spatial--temporal complexity.

\Cref{tab:tokens_groups} compares the token selection rates across the
three benchmarking groups.
The similar distributions confirm that the stratification preserves
distortion-type balance across groups.

\begin{table}[tb]
\centering
\caption{Problem token selection rates (\% of clips) per benchmarking
group. A token is selected when $\geq$2 assessors chose it.}
\label{tab:tokens_groups}
\begin{tabular}{@{}lrrr@{}}
\toprule
Token & TH & TH-BB & TH-BR \\
\midrule
Noise          & 62 & 70 & 55 \\
Low resolution & 45 & 50 & 50 \\
Lighting/color & 45 & 42 & 42 \\
No issue       & 22 & 28 & 40 \\
Blur           & 15 & 10 & 10 \\
Choppy motion  &  5 &  2 &  2 \\
Framerate      &  2 &  5 &  8 \\
Banding        &  5 &  8 &  8 \\
Other          &  0 &  5 &  5 \\
Blockiness     &  0 &  0 &  0 \\
\bottomrule
\end{tabular}
\end{table}


\section{Analysis}
\label{sec:analysis}

The full dataset of 838 clips (each 15\,s) can serve as training data
for data-driven models, while the stratified benchmarking subset of 120
clips (each trimmed to 10\,s) is designed for systematic evaluation.
We demonstrate the latter use case through codec compression efficiency analysis and a preliminary super-resolution evaluation.
We evaluate four encoders:
H.264~\cite{wiegand_overview_2003} (baseline, Intel Quick Sync Video,
hardware-accelerated),
H.265~\cite{sullivan_overview_2012} (Intel Quick Sync Video,
hardware-accelerated),
VVenC H.266~\cite{bross_overview_2021} (software), and
libaom AV1~\cite{han_technical_2021} (software, version 3.13.1).
All clips are encoded in a low-delay configuration at multiple fixed
quantization parameter (QP) levels
(QP\,=\,20--42 for H.264/H.265; adjusted ranges
for AV1 and H.266), following the methodology of
Naderi~\etal~\cite{naderi_vcd_2024};
exact encoder settings are listed in the supplementary material.
Bjontegaard delta rate
(BD-rate)~\cite{bjontegaard_calculation_2001} is computed per clip
relative to the H.264 baseline using Peak Signal-to-Noise Ratio (PSNR)
and Video Multi-Method Assessment Fusion
(VMAF)~\cite{li_vmaf_2018} as quality metrics.

\subsection{Comparison with Other Datasets}
\label{sec:comparison}

We compare BD-rate distributions across five datasets:
the HEVC common test sequences~\cite{bossen_common_2013},
UVG~\cite{mercat_uvg_2020},
VCD~\cite{naderi_vcd_2024} (desktop subset), the proposed
VCD2 benchmarking subset, and VCD2 after
VP8 pre-encoding (VCD2\,+\,VP8) which simulates WebRTC capture
conditions as in VCD.
\Cref{tab:bdrate_all} reports the mean BD-rate (\%) for each
dataset--encoder combination under both quality metrics.

\begin{table*}[tb]
\centering
\caption{Mean BD-rate (\%) relative to H.264 baseline.
95\% confidence intervals are shown in parentheses.
More negative values indicate greater bitrate savings.
VCD2\,+\,VP8 denotes VCD2 content pre-encoded with VP8 to simulate
WebRTC capture conditions.}
\label{tab:bdrate_all}
\small
\begin{tabular}{@{}lccccccc@{}}
\toprule
 &$N$ & \multicolumn{3}{c}{VMAF BD-Rate (\%)} & \multicolumn{3}{c}{PSNR BD-Rate (\%)} \\
\cmidrule(lr){3-5} \cmidrule(lr){6-8}
Dataset & Clips & H.265 & AV1 & H.266 & H.265 & AV1 & H.266 \\
\midrule
HEVC~\cite{bossen_common_2013}
  & 25 & $-34.9$ (3.0) & $-32.6$ (4.1) & $-65.0$ (3.9)
       & $-34.0$ (3.0) & $-36.1$ (5.1) & $-64.5$ (4.2) \\
UVG~\cite{mercat_uvg_2020}
  & 16 & $-34.5$ (5.2) & $-40.0$ (9.9) & $-70.4$ (7.5)
       & $-42.5$ (6.1) & $-48.2$ (9.4) & $-72.6$ (7.5) \\
VCD~\cite{naderi_vcd_2024}
  & 120 & $-27.9$ (2.3) & $-36.3$ (1.7) & $-63.7$ (1.8)
        & $-32.1$ (1.4) & $-40.0$ (1.8) & $-62.5$ (1.6) \\
VCD2 (ours)
  & 120 & $-25.9$ (2.7) & $-42.2$ (2.7) & $-71.3$ (1.7)
        & $-36.0$ (1.9) & $-52.8$ (2.4) & $-70.2$ (1.5) \\
VCD2\,+\,VP8
  & 120 & $-25.6$ (2.1) & $-35.1$ (1.4) & $-62.9$ (1.9)
        & $-35.0$ (1.3) & $-38.0$ (1.3) & $-61.4$ (1.5) \\
\bottomrule
\end{tabular}
\end{table*}

\begin{figure*}[tb]
\centering
\begin{subfigure}[b]{0.24\linewidth}
  \includegraphics[width=\linewidth]{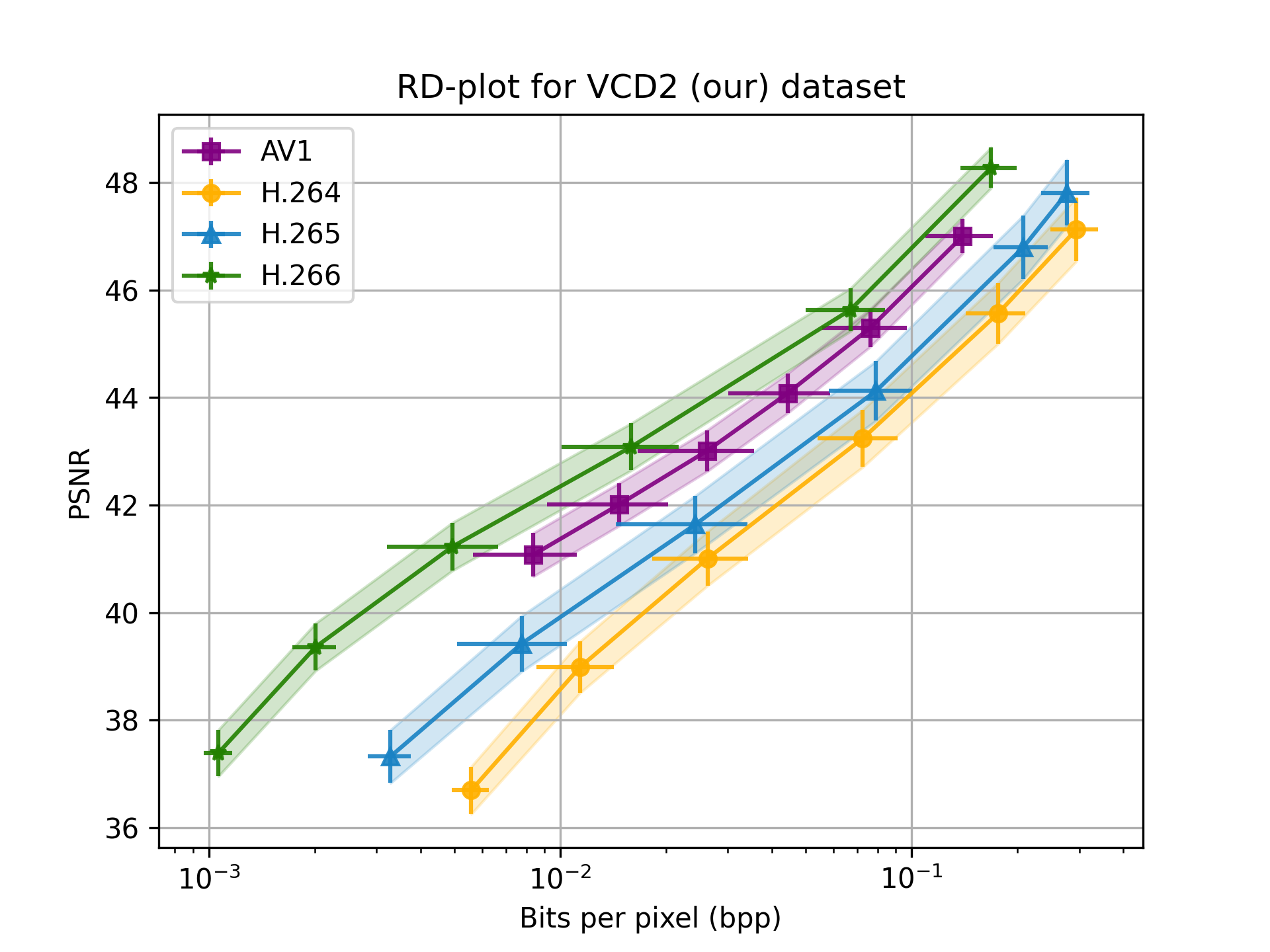}
  \caption{VCD2, PSNR}
  \label{fig:rd_psnr}
\end{subfigure}
\hfill
\begin{subfigure}[b]{0.24\linewidth}
  \includegraphics[width=\linewidth]{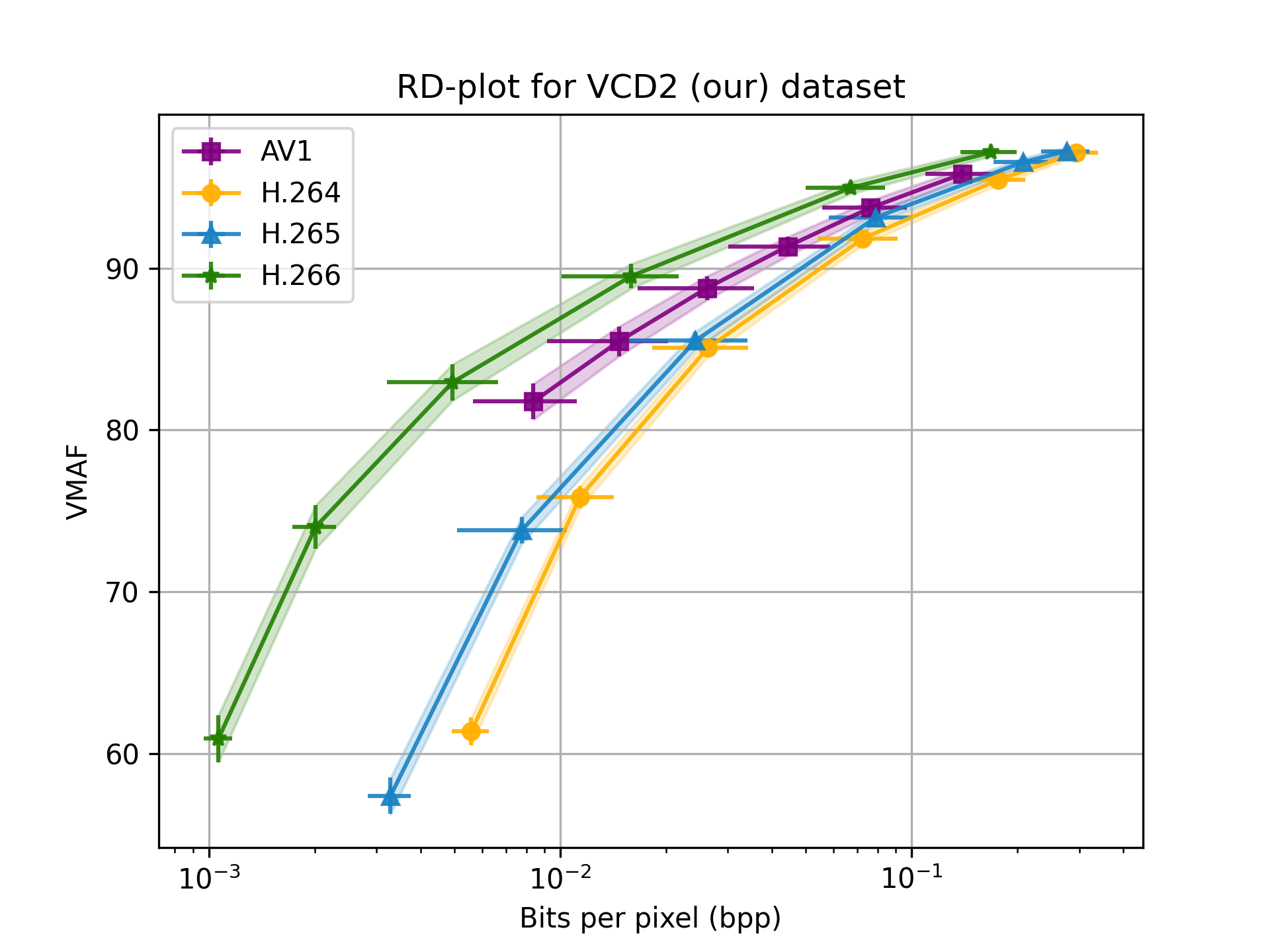}
  \caption{VCD2, VMAF}
  \label{fig:rd_vmaf}
\end{subfigure}
\hfill
\begin{subfigure}[b]{0.24\linewidth}
  \includegraphics[width=\linewidth]{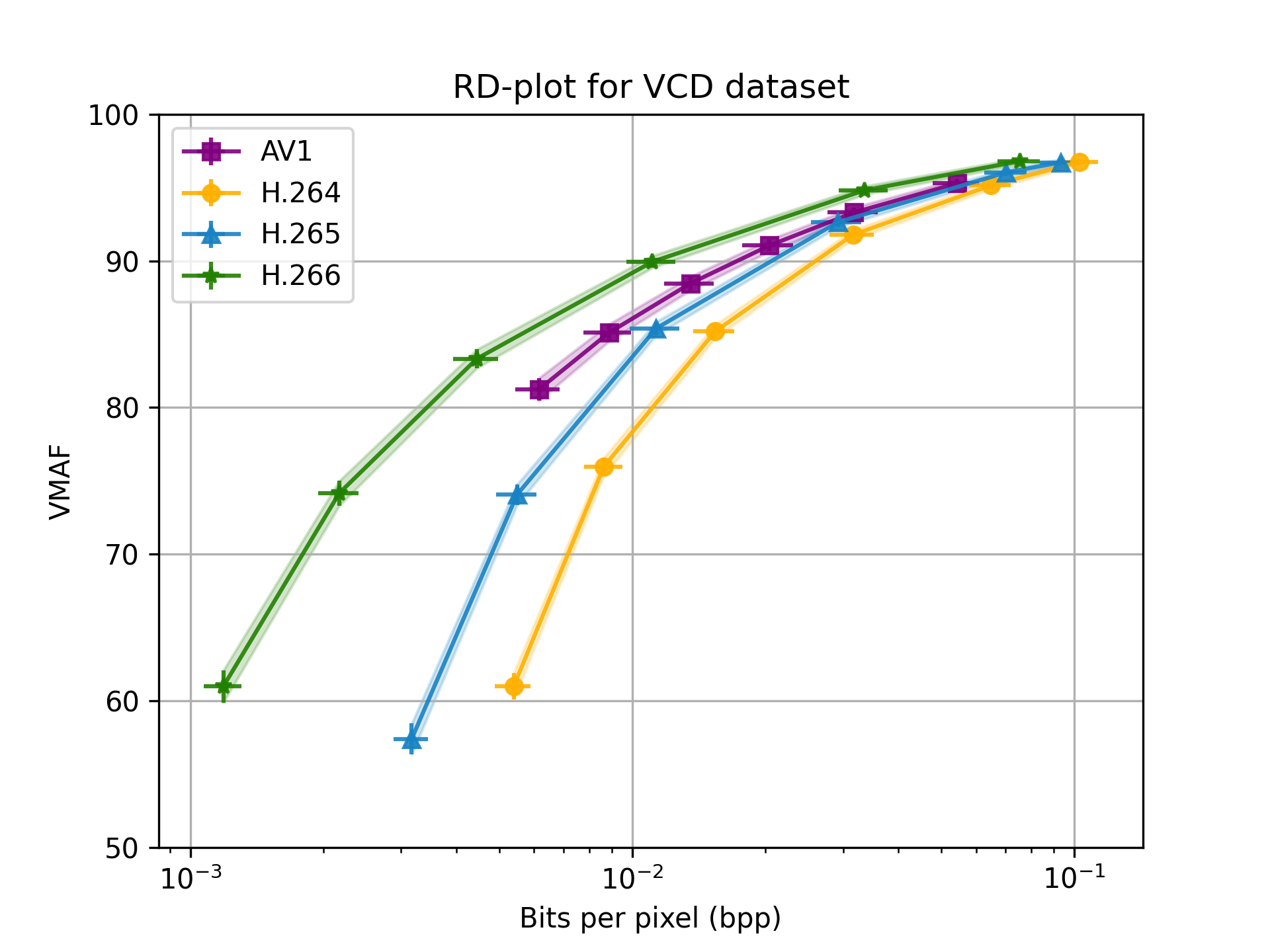}
  \caption{VCD, VMAF}
  \label{fig:rd_vmaf2}
\end{subfigure}
\hfill
\begin{subfigure}[b]{0.24\linewidth}
  \includegraphics[width=\linewidth]{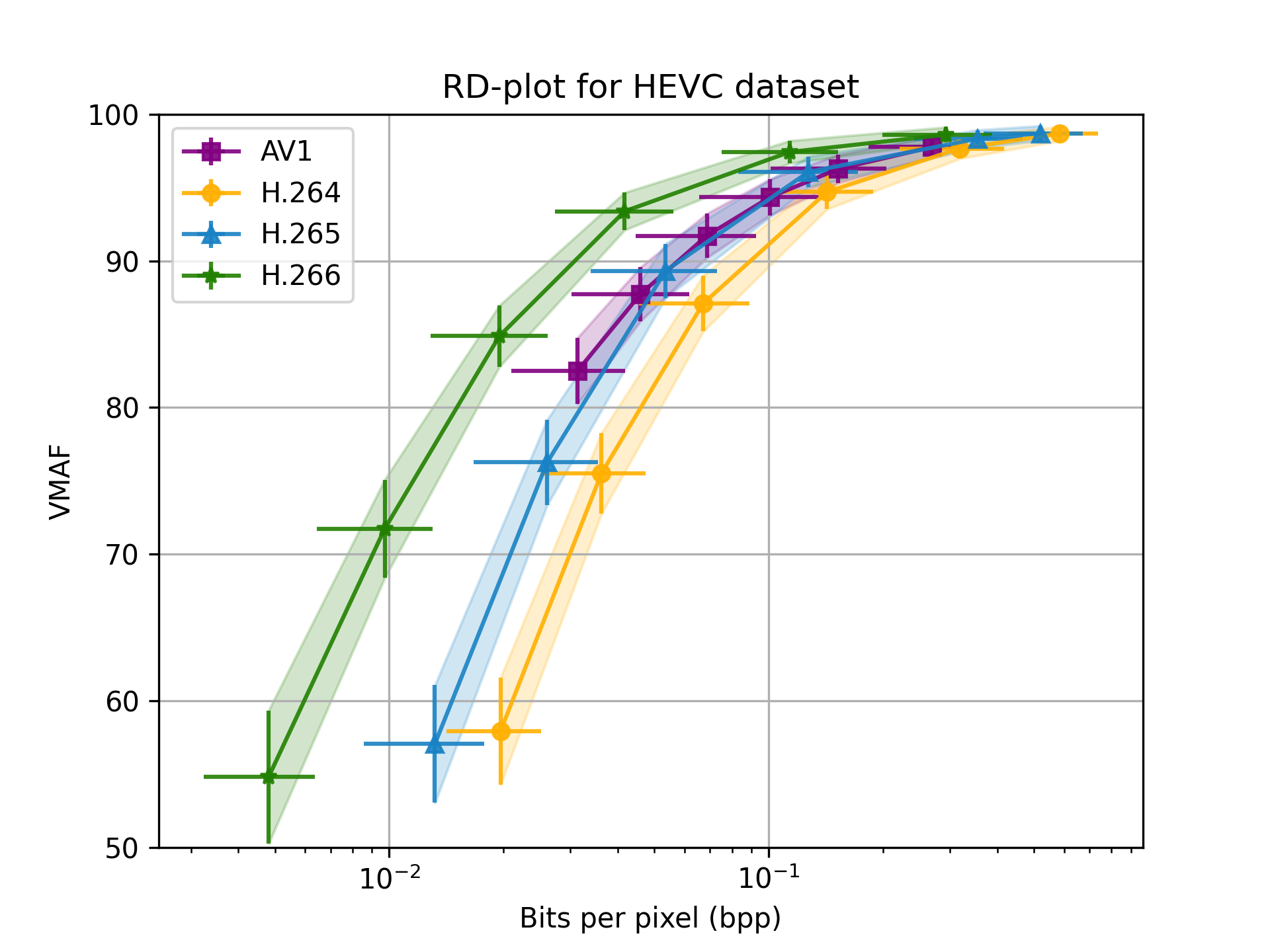}
  \caption{HEVC, VMAF}
  \label{fig:rd_vmaf3}
\end{subfigure}
\caption{Rate--distortion curves for the VCD2 benchmarking subset
(\subref{fig:rd_psnr}--\subref{fig:rd_vmaf}) and VMAF rate--distortion
curves for VCD (\subref{fig:rd_vmaf2}) and HEVC
(\subref{fig:rd_vmaf3}) datasets.
Shaded bands indicate 95\% confidence intervals.}
\label{fig:rd_plots}
\end{figure*}

\Cref{fig:rd_plots} shows the rate--distortion curves for the VCD2
benchmarking subset averaged across clips, with 95\% confidence bands.
H.266 and AV1 consistently outperform H.264 and H.265 across the entire
bitrate range under both metrics, with H.266 achieving the largest gains
at low bitrates.

A two-way mixed analysis of variance (ANOVA) with encoder
(within-subject) and dataset (between-subject) on VMAF BD-rate
($N = 255$ clips with complete BD-rate estimates across all encoders)
reveals a significant main effect of encoder, $F(2, 502) = 937.10$,
$p < .001$, $\eta_p^2 = .789$\footnote{Greenhouse--Geisser-corrected $p$-values
are reported where sphericity was violated}, confirming that codec
generation is the dominant factor in compression efficiency.
The main effect of dataset is also significant,
$F(3, 251) = 2.75$, $p = .044$, $\eta_p^2 = .032$.
A significant encoder$\times$dataset interaction,
$F(6, 502) = 10.59$, $p < .001$, $\eta_p^2 = .112$,
indicates that the relative advantage of encoders depends on the content.
PSNR-based results confirm these patterns with a larger dataset effect
($F(3, 251) = 16.80$, $p < .001$, $\eta_p^2 = .167$).

As shown in \cref{tab:bdrate_all}, the VCD2 subset yields larger
BD-rate savings for software codecs (AV1 and H.266) than VCD, consistent
with the significant encoder$\times$dataset interaction.
The PSNR and VMAF metrics agree for AV1 and H.266 but diverge for H.265,
where the VMAF BD-rate on VCD2 shows reduced savings compared to VCD
while the PSNR BD-rate is comparable, suggesting that the choice of
quality metric affects the measured compression advantage of hardware
H.265 on webcam content.

\subsection{Effect of Background Processing}
\label{sec:background_effect}

Having established that content type influences codec efficiency across
datasets, we next examine whether background processing within the
VCD2 subset produces similar effects.
We refer to the three benchmarking groups (TH, TH-BB, TH-BR) as
content conditions in the statistical analyses that follow.
A two-way mixed ANOVA with encoder (within-subject) and content
condition (between-subject) on VMAF BD-rate reveals
significant main effects of both encoder ($F(2, 202) = 344.74$,
$p < .001$, $\eta_p^2 = .773$) and content condition
($F(2, 101) = 29.64$, $p < .001$, $\eta_p^2 = .370$), as well as a
significant interaction ($F(4, 202) = 8.84$, $p < .001$,
$\eta_p^2 = .149$).
The content condition effect is larger under VMAF ($\eta_p^2 = .370$)
than under PSNR ($\eta_p^2 = .185$), indicating that background
processing has a stronger influence on perceptual-quality-based
compression efficiency.

Background replacement (TH-BR) yields the largest BD-rate savings for
AV1 ($-58.9\%$) and H.266 ($-77.1\%$), compared to $-30.0\%$ and
$-71.6\%$ on original content (TH), respectively.
H.265 shows a narrower range across conditions ($-24.3\%$ to $-29.5\%$)
than AV1, though wider than H.266.
The PSNR-based interaction is notably larger ($\eta_p^2 = .479$),
with AV1 on TH-BR reaching $-66.5\%$ compared to $-45.4\%$ on TH.
These results indicate that the simplified backgrounds in TH-BR
content are more efficiently exploited by modern software codecs,
while hardware H.265 does not benefit to the same extent.
Per-condition BD-rate statistics are provided in the supplementary
material.

\subsection{Effect of Perceptual Distortions}
\label{sec:distortion_effect}

We investigate whether codec compression efficiency is affected by the
presence of perceptual distortions captured by the quality tokens
(\cref{sec:annotation}).
For each of the eight tokens with sufficient variance (excluding
blockiness and choppy motion), a linear mixed-effects model (LMM) is
fitted with BD-rate as the response, encoder, content condition, and
the proportion of assessors selecting the token as fixed effects
(including all interactions), and source clip as a random intercept.

Under PSNR, no token shows a statistically significant effect on
BD-rate ($p > .05$ for all terms), suggesting that PSNR-based
compression efficiency is not substantially associated with the
perceptual distortions captured by the tokens.

Under VMAF, the noise token shows a significant encoder-specific
effect.
For H.265, higher noise prevalence is associated with worse
(\ie, less negative) VMAF BD-rate on original content
(coefficient $= +40.4$, $p < .001$), meaning that the rate--distortion
advantage of H.265 over H.264 diminishes on noisy clips.
This penalty is attenuated in TH-BB ($p = .006$) and TH-BR
($p = .020$), suggesting that background processing partially mitigates
the noise-related efficiency loss---which is expected, as background
blur and replacement affect the majority of the picture area and
thereby reduce the spatial noise that the encoder must code.
Neither AV1 nor H.266 shows significant sensitivity to the noise token,
indicating that these codecs maintain a stable performance ratio
relative to H.264 across noise levels---that is, both the test codec
and the baseline appear similarly affected by noise, preserving their
relative efficiency gap.
The no-issue token also shows a significant effect for H.265:
clips perceived as artifact-free yield better VMAF BD-rate
(coefficient $= -35.9$, $p = .002$).

These findings indicate that perceptual distortions, particularly noise,
can influence codec evaluation outcomes in metric- and
encoder-dependent ways, highlighting the importance of including
realistic source-level distortions in benchmarking content.

\subsection{Effect of Lossy Capture Compression}
\label{sec:vp8_effect}

Beyond source-level distortions, the recording pipeline itself can
alter codec evaluation outcomes.
To quantify this effect, we simulate a WebRTC-style capture pipeline by
pre-encoding the VCD2 benchmarking clips with VP8 at a constant
bitrate of 2500\,kbps using the real-time preset
(see supplementary material) and then repeating the codec efficiency
evaluation on the decoded output.
This setup approximates the recording conditions of the
VCD dataset~\cite{naderi_vcd_2024}, which was captured through a
WebRTC-based application that uses VP8 or H.264
encoding~\cite{alvestrand_webrtc_2016}.

A two-way repeated measures ANOVA ($N = 117$ clips) reveals a
significant main effect of VP8 pre-processing on VMAF BD-rate
($F(1, 116) = 113.10$, $p < .001$, $\eta_p^2 = .494$) and a
significant VP8$\times$encoder interaction ($F(2, 232) = 27.42$,
$p < .001$, $\eta_p^2 = .191$)\footnote{Greenhouse--Geisser-corrected
$p$-values are reported where Mauchly's test indicated sphericity
violations.}, indicating that the degradation is encoder-dependent.
Post-hoc paired comparisons (Bonferroni-corrected) show that VP8
pre-processing significantly increases VMAF BD-rate for AV1
($+7.1$\,pp, $d = 0.60$, $p < .001$) and H.266 ($+7.7$\,pp,
$d = 1.36$, $p < .001$), while H.265 remains unaffected
($+0.3$\,pp, $p_\text{Bonf} = 1.00$).
The overall VMAF BD-rate worsens by 5.0\,pp ($d = 0.98$); PSNR-based
results confirm the same pattern with larger effect sizes
($\eta_p^2 = .760$ for the main effect; overall $+8.0$\,pp,
$d = 1.77$).

As shown in \cref{tab:bdrate_all}, the VCD2\,+\,VP8 BD-rates closely
resemble those of VCD, consistent with the fact that VCD was recorded
through a WebRTC pipeline.
The encoder efficiency ranking (H.266, AV1, H.265, from most to least
efficient) is preserved regardless of VP8 pre-processing.
These results indicate that lossy capture compression reduces the
content-dependent signal variation that advanced codecs exploit,
diminishing their measured advantage over simpler encoders.
Lossless references are therefore important for accurately
characterizing the efficiency gains of modern codecs on webcam content.

\subsection{Source Format Robustness}
\label{sec:source_format}

Whether the camera output format---MJPEG versus uncompressed
YUYV422---affects codec evaluation conclusions is tested by comparing
BD-rates between YUYV422 ($N = 11$) and MJPEG ($N = 29$) clips within
TH using a mixed ANOVA.
Source format has no significant effect
(PSNR: $p = .785$, $\eta_p^2 = .002$;
VMAF: $p = .182$, $\eta_p^2 = .046$) and no interaction with encoder
(PSNR: $p = .593$; VMAF: $p = .553$), supporting the use of all clips
as a unified evaluation corpus (full results in supplementary material).

\subsection{Super-Resolution Evaluation}
\label{sec:sr_eval}

To evaluate applicability beyond codec benchmarking, four SR
models---RealSR~\cite{ji2020realsr}, SwinIR~\cite{liang2021swinir},
RealViformer~\cite{wu2024realviformer}, and
BasicVSR++~\cite{chan_basicvsr_2022}---are applied to clips downscaled
to 180p and upscaled to 720p ($4\times$).
We evaluate on four dataset groups: VCD2, VCD2\,+\,VP8,
REDS~\cite{nah_ntire_2019}, and VCD~\cite{naderi_vcd_2024}.

\begin{table}[tb]
\centering
\caption{Mean VMAF after $4\times$ SR (180p $\to$ 720p) with 95\% CI.
Best model per dataset in bold.}
\label{tab:sr_results}
\small
\begin{tabular}{@{}lcccc@{}}
\toprule
Dataset & RealSR & SwinIR & RealVif. & BVSR++ \\
\midrule
REDS
  & $61.4$ {\scriptsize(3.6)}
  & $\mathbf{72.6}$ {\scriptsize(2.7)}
  & $37.5$ {\scriptsize(6.2)}
  & $71.5$ {\scriptsize(2.8)} \\
VCD
  & $72.6$ {\scriptsize(3.5)}
  & $\mathbf{78.6}$ {\scriptsize(3.1)}
  & $51.9$ {\scriptsize(4.6)}
  & $77.9$ {\scriptsize(3.0)} \\
VCD2 (ours)
  & $79.7$ {\scriptsize(1.5)}
  & $\mathbf{90.9}$ {\scriptsize(1.2)}
  & $58.8$ {\scriptsize(2.5)}
  & $90.1$ {\scriptsize(1.4)} \\
VCD2\,+\,VP8
  & $77.1$ {\scriptsize(1.6)}
  & $\mathbf{84.0}$ {\scriptsize(1.5)}
  & $55.7$ {\scriptsize(2.7)}
  & $83.7$ {\scriptsize(1.6)} \\
\bottomrule
\end{tabular}
\end{table}

\Cref{tab:sr_results} reports mean VMAF scores per model and dataset.
A two-way ANOVA reveals a significant dataset
effect on absolute scores
(VMAF: $\eta_p^2 = .094$; PSNR: $\eta_p^2 = .144$;
SSIM: $\eta_p^2 = .230$; all $p < .001$),
while the Model$\times$Dataset interaction
is not significant ($p = .115$--$.628$, $\eta_p^2 \approx .005$).
SwinIR and BasicVSR++ consistently outperform RealSR and RealViformer
across all datasets.
VCD2 yields the highest absolute scores, reflecting camera-native
signal quality; VP8 pre-processing reduces VMAF by 3--7 points.
The dataset is thus applicable for SR model benchmarking, while the
significant dataset main effect underscores the importance of evaluation
corpus choice.


\section{Conclusion}
\label{sec:conclusion}

We presented a camera-native talking-head webcam video dataset comprising 838
losslessly encoded recordings from 799 participants across 443 camera-code
categories.
The dataset preserves camera-driver output through FFV1 lossless encoding,
eliminating double-compression artifacts present in conventional webcam
recording workflows.
Each recording is annotated with MOS and ten perceptual quality tokens.
A stratified benchmarking subset of 120 clips in three content conditions
(TH, TH-BB, TH-BR) enables systematic evaluation.

Codec analysis across four datasets revealed significant
encoder$\times$dataset ($\eta_p^2 = .112$) and
encoder$\times$content condition ($\eta_p^2 = .149$) interactions under
VMAF, demonstrating that both content type and background processing
affect compression efficiency.
Simulating WebRTC-style capture by pre-encoding with VP8 confirmed that
platform-level compression reduces the measured BD-rate advantage of AV1
and H.266, with the resulting BD-rates aligning closely with VCD.
A preliminary SR evaluation showed that the choice of evaluation dataset
significantly affects absolute SR performance while preserving model
rankings.
These findings collectively indicate that source-reference
construction---including capture-pipeline compression---is not a neutral
preprocessing detail but an experimental variable affecting benchmark
conclusions.

Beyond codec and SR evaluation, the dataset supports potential
applications in video quality assessment, restoration, and denoising,
where real-life webcam video with authentic camera-sensor artifacts is
essential;
69.8\% of clips exhibit noise and 36.6\% lighting/color
issues, providing naturally occurring degradation patterns for these tasks.

\textbf{Limitations.}
The dataset targets desktop/laptop webcam RTC; mobile devices are not
included.
Camera ISP processing is part of the captured signal, reflecting what
real-world applications receive but not raw sensor data.
The TH-BR condition uses four virtual backgrounds; real-world usage
encompasses greater diversity.

The published videos, benchmarking subset, MOS ratings, and perceptual
quality tokens are open-sourced at \url{https://github.com/microsoft/VCD}.

\bibliographystyle{IEEEtran}
\bibliography{main,IC3-AI}

\appendices

\section{Data Collection Details}
\label{sec:supp_data_collection}

This section provides additional technical details on the recording
pipeline summarized in Sec. 2.1.

\paragraph{Pixel format selection.}
Consumer webcams expose multiple output formats over the USB Video Class
(UVC) protocol, and the choice of format determines signal fidelity before
any software processing occurs.
The recording application implements a hierarchical pixel format priority:
(1)~YUYV422 (uncompressed 4:2:2 chroma subsampling, 16~bits/pixel),
(2)~NV12 (uncompressed 4:2:0, 12~bits/pixel), and
(3)~MJPEG (Motion JPEG, lossy intra-frame compression) as a fallback.
Uncompressed formats preserve the camera's decoded sensor signal without
additional quantization, whereas MJPEG introduces DCT-based compression
artifacts within the camera firmware before the signal reaches the host.

For YUYV422 and NV12 inputs, the software performs only a lossless memory
layout conversion (packed to planar format) before encoding.
For MJPEG inputs, the software decodes the JPEG stream and preserves the
detected color range (full or limited) through metadata tagging;
these clips are losslessly stored after camera compression, meaning
their quality ceiling is bounded by the MJPEG encoding applied in camera
firmware.

\paragraph{Lossless encoding.}
All recordings are encoded with the FFV1 lossless video codec (Level~3)
in a Matroska (.mkv) container.
FFV1 uses context-adaptive arithmetic coding to achieve lossless
intra-frame compression, reducing file size relative to uncompressed
storage while guaranteeing bit-exact reconstruction of the input signal.
We configure FFV1 Level~3 encoding with 16 slices per frame and per-slice
CRC-32 integrity checks, allowing detection of bit corruption during
storage and transmission.
The codec guarantees bit-exact reconstruction: decoded pixel values are
identical to the values provided to the encoder.

\paragraph{Frame timing and buffering.}
The recording application operates in passthrough frame timing mode,
which preserves the original frame timestamps from the camera without
interpolation, frame dropping, or duplicate frame insertion.
A 256\,MB ring buffer is allocated to absorb I/O latency spikes and
prevent frame drops during disk writes.

\paragraph{Signal fidelity.}
Before data reaches the recording software, the firmware of consumer UVC
webcams typically applies demosaicing, white balance, gamma correction
(sRGB), noise reduction, and auto-exposure adjustments.
For the webcams used in this dataset, these processing steps are applied
in camera firmware and are not bypassable through the standard UVC
capture interface.
The recording pipeline adds no further lossy processing: YUYV422 and
NV12 inputs undergo only lossless format conversion, and all frames are
encoded with the mathematically lossless FFV1 codec.
This approach avoids adding a second lossy compression stage after
camera output and therefore prevents the double-compression artifacts
(\ie camera compression followed by capture-software compression)
present in conventional webcam recording workflows.

\section{Full Dataset Statistics}
\label{sec:supp_dataset_stats}

\Cref{tab:supp_dataset_stats} provides summary dataset statistics,
including per-scenario clip counts, resolution and pixel format
distributions, and per-recording self-reported demographics.

\begin{table}[tb]
\centering
\caption{Full statistics of the 838-clip published dataset.}
\label{tab:supp_dataset_stats}
\small
\begin{tabular}{@{}p{0.57\columnwidth}p{0.35\columnwidth}@{}}
\toprule
Property & Value \\
\midrule
Published clips & 838 \\
Unique participants & 799 \\
Camera-code categories\textsuperscript{$\dagger$} & 443 \\
Duration per clip & 15\,s \\
Codec & FFV1 (lossless) \\
Container & Matroska (.mkv) \\
\midrule
\multicolumn{2}{@{}l}{\textit{Resolution distribution}} \\
\quad 1280$\times$720 (720p) & 509 (60.7\%) \\
\quad 1920$\times$1080 (1080p) & 277 (33.1\%) \\
\quad 2560$\times$1440 (1440p) & 43 (5.1\%) \\
\quad 3840$\times$2160 (4K) & 8 (1.0\%) \\
\quad 1920$\times$1440 & 1 (0.1\%) \\
\midrule
\multicolumn{2}{@{}l}{\textit{Input pixel format}} \\
\quad MJPEG (camera-compressed) & 631 (75.3\%) \\
\quad YUYV422 (uncompressed) & 206 (24.6\%) \\
\quad NV12 (uncompressed) & 1 (0.1\%) \\
\midrule
\multicolumn{2}{@{}l}{\textit{Recording scenarios}} \\
\quad S01: Continuous slow body movement & 314 (37.5\%) \\
\quad S02: Hand counting exercise & 154 (18.4\%) \\
\quad S03: Text reading exercise & 300 (35.8\%) \\
\quad S04: Natural video call behavior & 70 (8.4\%) \\
\midrule
\multicolumn{2}{@{}l}{\textit{Per-recording self-reported demographics}} \\
\quad Female / Male / Prefer not to say & 310 / 527 / 1 \\
\quad White / Black / Asian / Mixed / Other & 570 / 145 / 64 / 48 / 11 \\
\midrule
MOS range (ACR, 5-point) & 1.00--5.00 (mean 3.35) \\
\bottomrule
\multicolumn{2}{@{}p{0.92\columnwidth}@{}}{\footnotesize\textsuperscript{$\dagger$}442 observed USB vendor/product codes plus one category for unavailable codes. The 838 clips come from 799 participants; 39 contributed two recordings.} \\
\end{tabular}
\end{table}

\section{Per-Group BD-Rate Statistics}
\label{sec:supp_groups}

This section reports the mean BD-rate (\%) for each encoder--group
combination in the NR-THD benchmarking subset: original talking-head
(TH), talking-head with background blur (TH-BB), and talking-head with
background replacement (TH-BR).
Values in \cref{tab:supp_groups} are computed relative to the H.264
baseline.

\begin{table}[tb]
\centering
\caption{Mean BD-rate (\%) relative to H.264 baseline per benchmarking
group, with 95\% $t$-based confidence intervals computed over clips
within each group.
VMAF and PSNR metrics are reported separately.}
\label{tab:supp_groups}
\begin{tabular}{@{}llrrr@{}}
\toprule
Metric & Group & H.265 & AV1 & H.266 \\
\midrule
\multirow{3}{*}{VMAF}
  & TH    & $-24.3 \pm 5.0$ & $-30.0 \pm 3.2$ & $-71.6 \pm 2.4$ \\
  & TH-BB & $-24.0 \pm 4.5$ & $-37.9 \pm 1.6$ & $-65.2 \pm 3.3$ \\
  & TH-BR & $-29.5 \pm 4.6$ & $-58.9 \pm 3.1$ & $-77.1 \pm 2.0$ \\
\midrule
\multirow{3}{*}{PSNR}
  & TH    & $-39.4 \pm 3.4$ & $-45.4 \pm 3.2$ & $-69.5 \pm 2.9$ \\
  & TH-BB & $-39.1 \pm 2.3$ & $-46.3 \pm 2.2$ & $-66.5 \pm 2.4$ \\
  & TH-BR & $-29.4 \pm 3.2$ & $-66.5 \pm 2.7$ & $-74.6 \pm 1.6$ \\
\bottomrule
\end{tabular}
\end{table}

\begin{figure*}[tb]
\centering
\begin{subfigure}[b]{0.32\linewidth}
  \includegraphics[width=\linewidth]{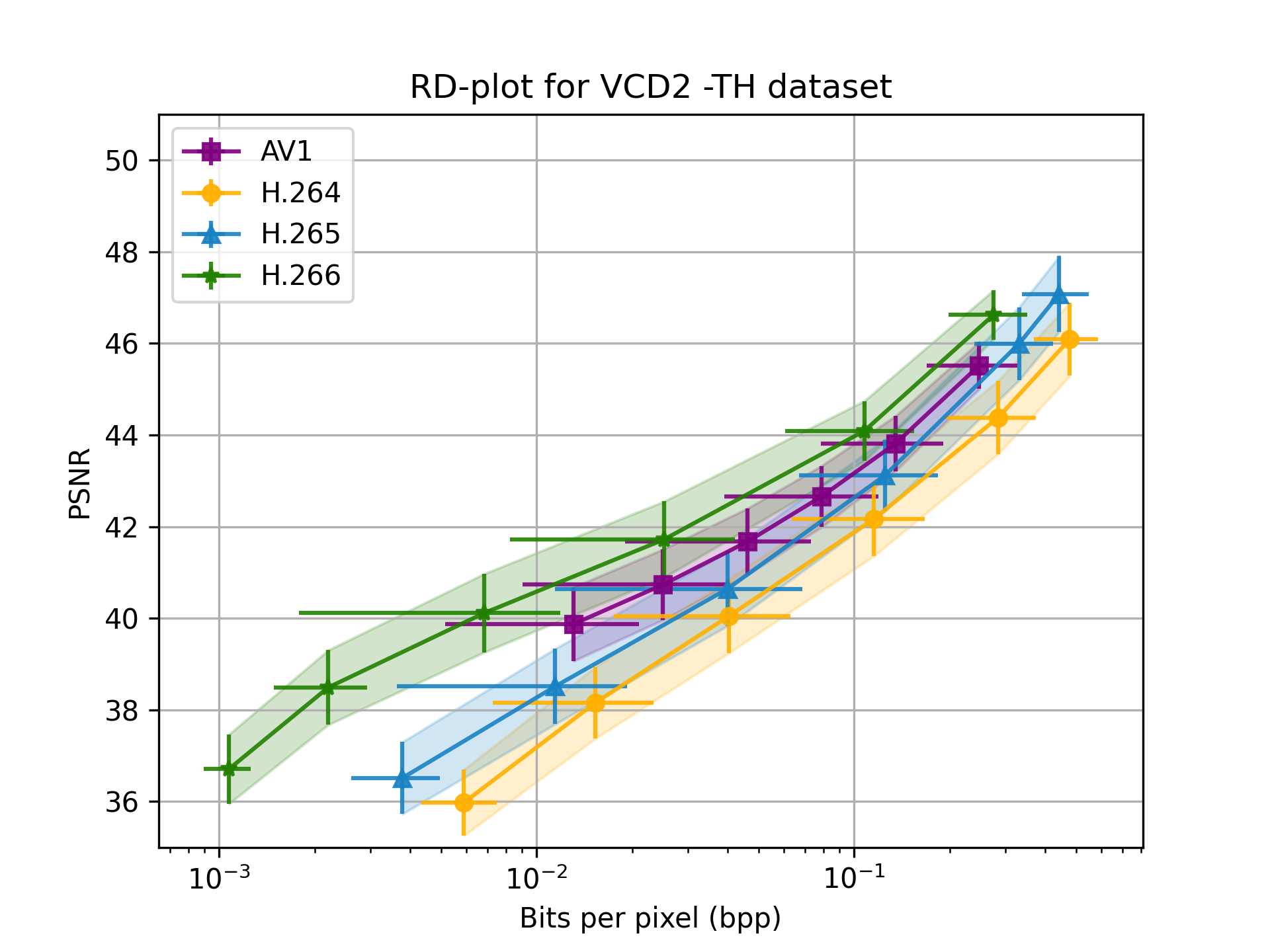}
  \caption{TH, PSNR}
  \label{fig:supp_rd_th_psnr}
\end{subfigure}
\hfill
\begin{subfigure}[b]{0.32\linewidth}
  \includegraphics[width=\linewidth]{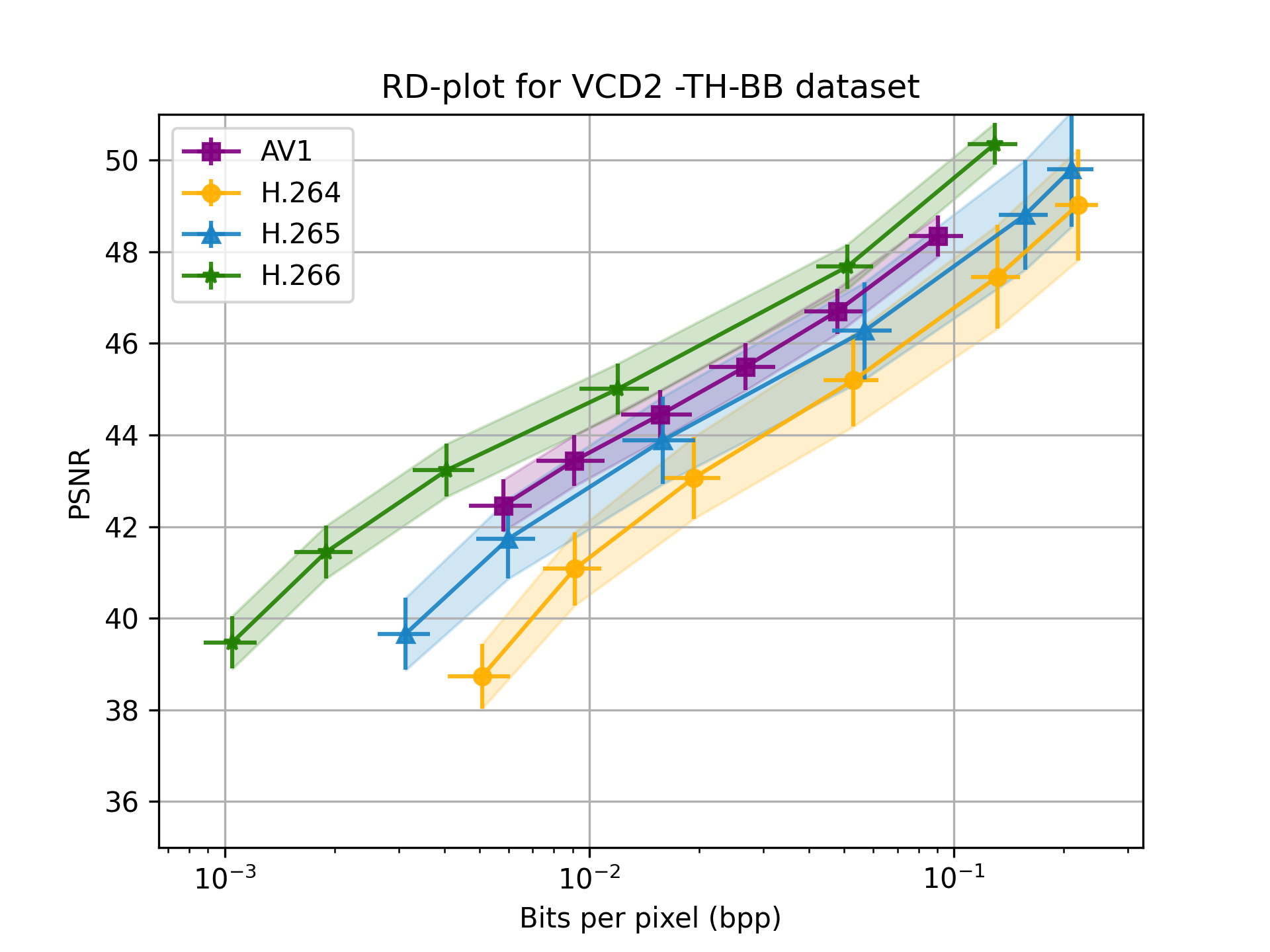}
  \caption{TH-BB, PSNR}
  \label{fig:supp_rd_thbb_psnr}
\end{subfigure}
\hfill
\begin{subfigure}[b]{0.32\linewidth}
  \includegraphics[width=\linewidth]{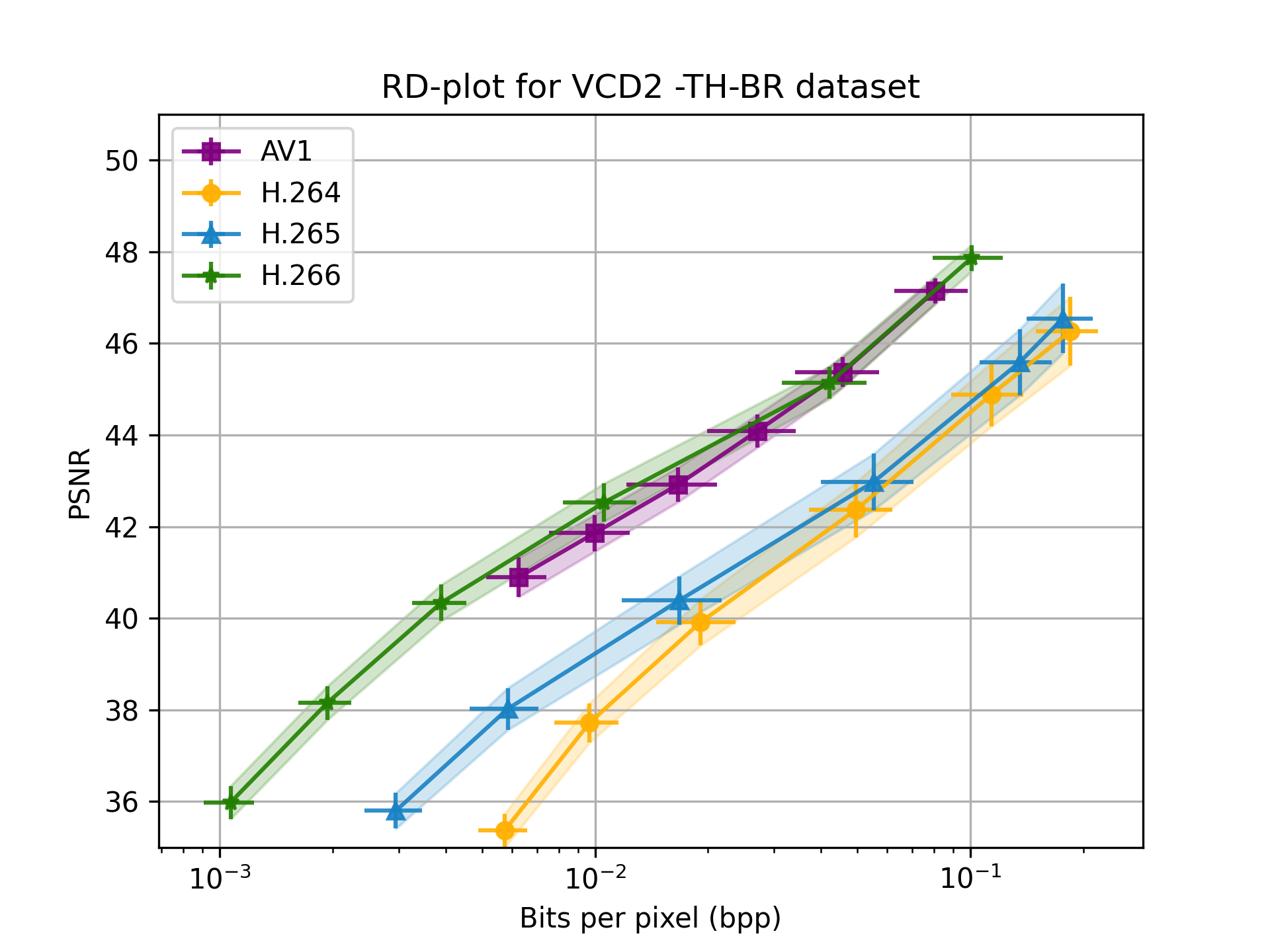}
  \caption{TH-BR, PSNR}
  \label{fig:supp_rd_thbr_psnr}
\end{subfigure}

\medskip

\begin{subfigure}[b]{0.32\linewidth}
  \includegraphics[width=\linewidth]{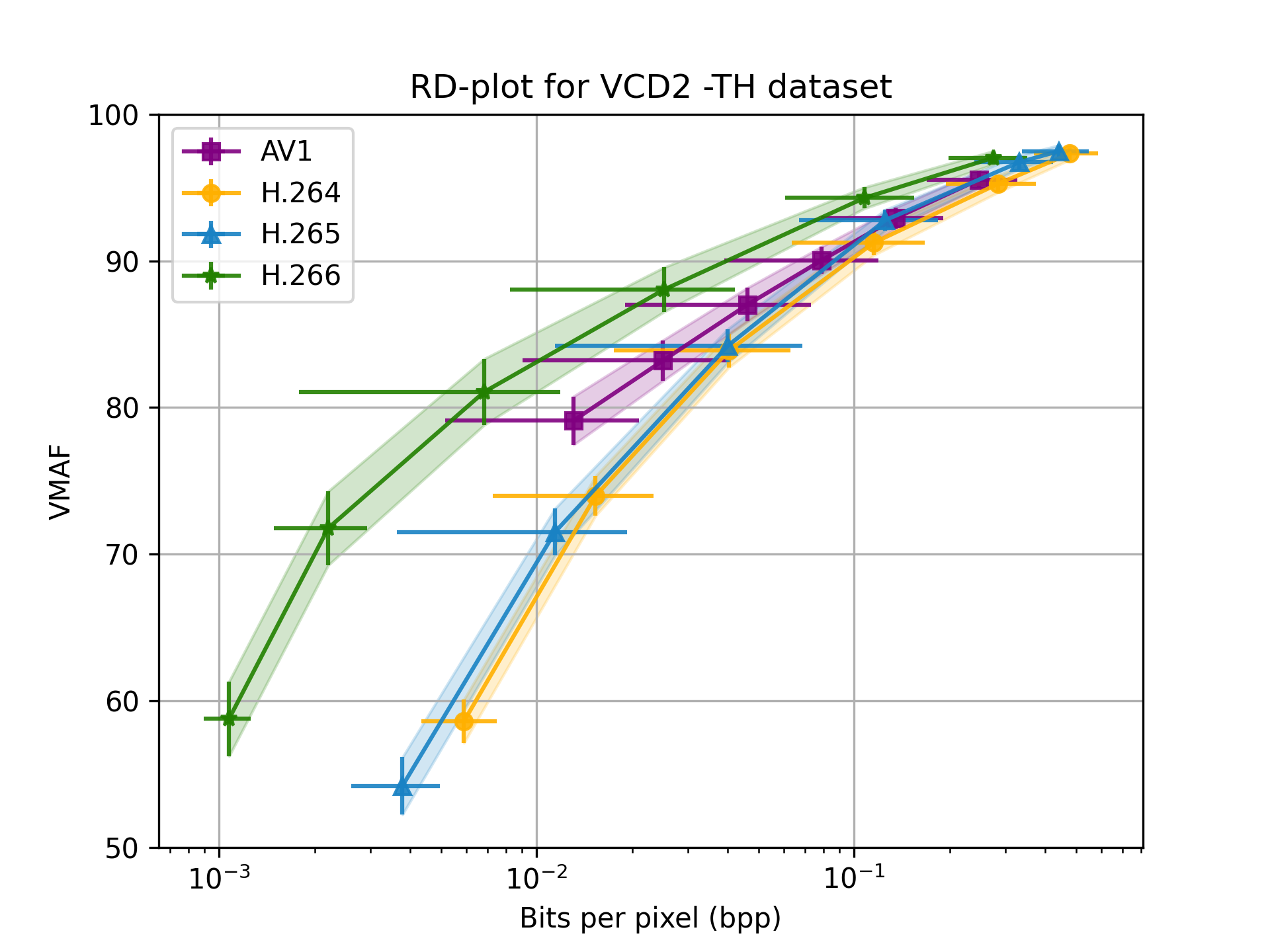}
  \caption{TH, VMAF}
  \label{fig:supp_rd_th_vmaf}
\end{subfigure}
\hfill
\begin{subfigure}[b]{0.32\linewidth}
  \includegraphics[width=\linewidth]{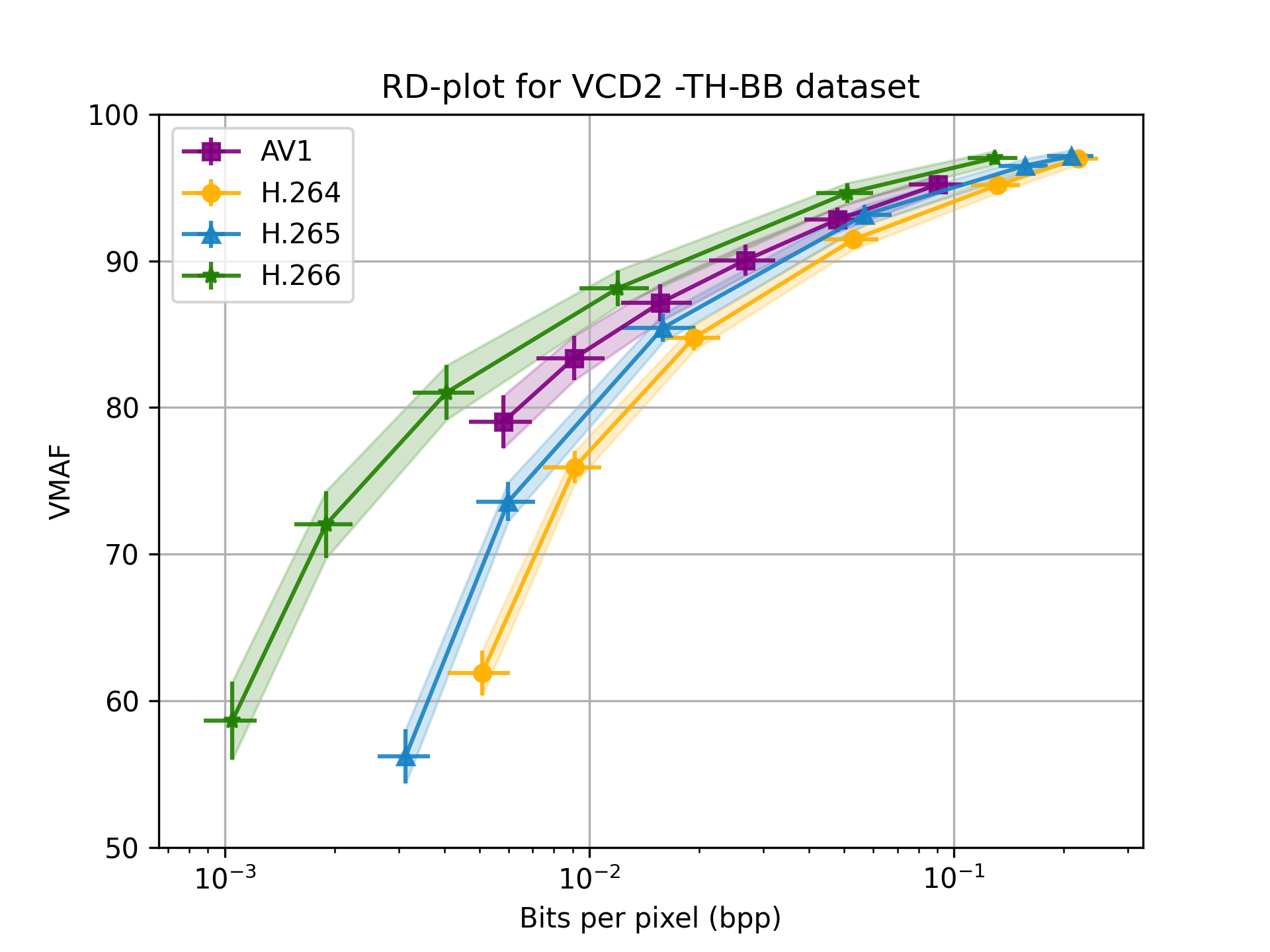}
  \caption{TH-BB, VMAF}
  \label{fig:supp_rd_thbb_vmaf}
\end{subfigure}
\hfill
\begin{subfigure}[b]{0.32\linewidth}
  \includegraphics[width=\linewidth]{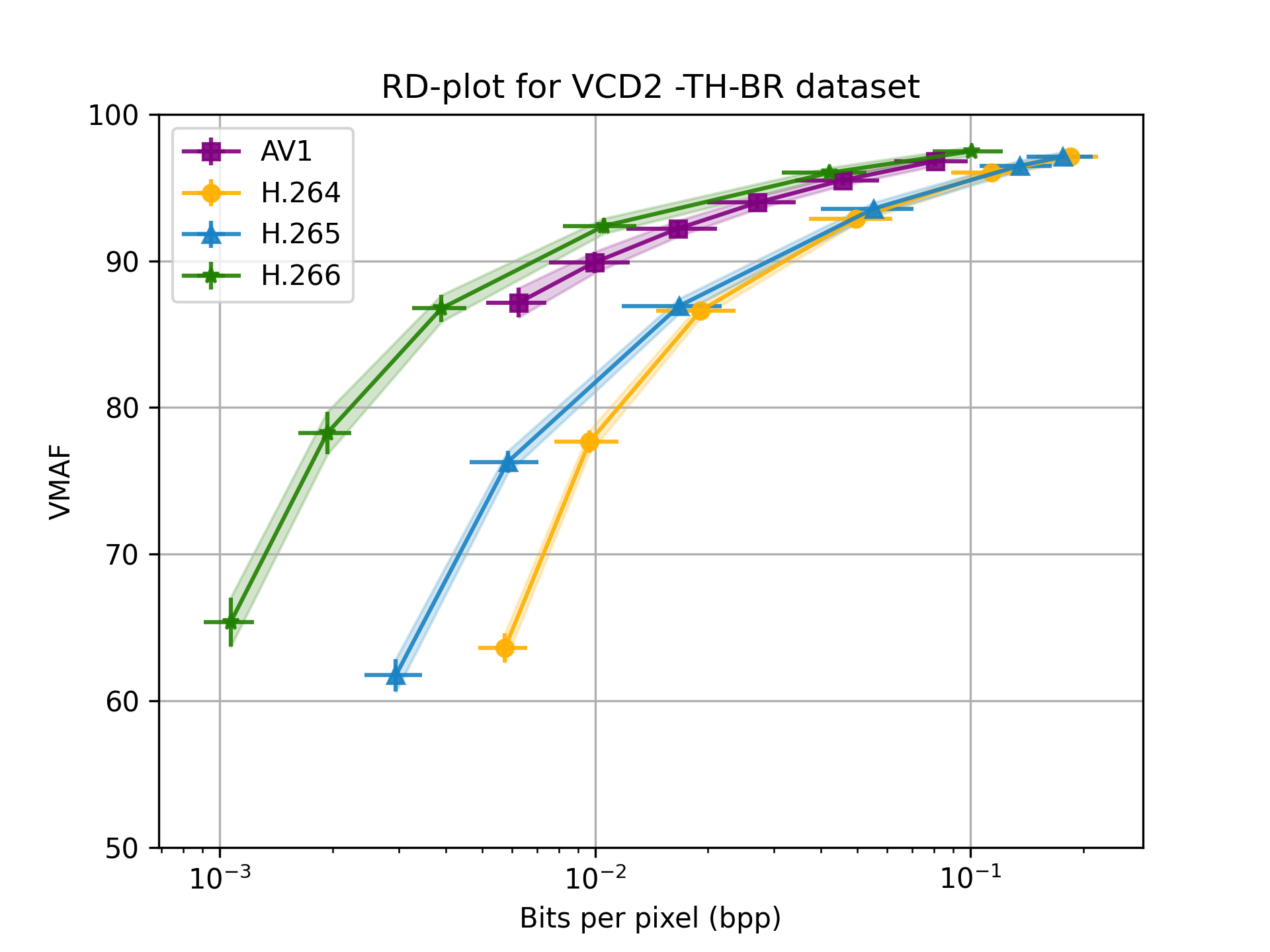}
  \caption{TH-BR, VMAF}
  \label{fig:supp_rd_thbr_vmaf}
\end{subfigure}
\caption{Rate--distortion curves per benchmarking group.
Top row: PSNR vs.\ bpp; bottom row: VMAF vs.\ bpp.
Each curve shows the mean metric value across clips in the group.}
\label{fig:supp_rd_groups}
\end{figure*}

\Cref{fig:supp_rd_groups} shows the rate--distortion curves for each
benchmarking group, plotting mean PSNR and VMAF against bits per pixel
(bpp) across all four codecs.

\section{Encoding Configuration}
\label{sec:supp_encoding}

\Cref{tab:supp_encoding} lists the exact command lines used in the
codec benchmarking experiments.
All encoders are configured for low-delay operation (no B-frames, no
picture reordering) with a single encoding pass and single-threaded
execution to ensure deterministic output.
Compression is controlled via fixed quantization parameter (QP) sweeps;
no rate-control mode is used.
The resolution and frame rate flags in the H.266 command are set per
clip; the values shown are representative.

Hardware-accelerated H.264 and H.265 encoding was performed on an Intel
Core i7-13800H (14 cores) with Intel Iris Xe Graphics (PCI device ID
8086:A7A0), running Windows~11 Enterprise 25H2, Intel graphics driver
32.0.101.673, and FFmpeg build 2026-02-18-git-52b676bb2.
AV1 encoding used libaom (AOMedia Project AV1 Encoder) version 3.13.1;
the \texttt{--i422} or \texttt{--i420} flag was selected depending on
the input pixel format.
H.266 encoding used VVenC version 1.14.0 (64-bit, SIMD=AVX2).
All inputs were converted to YUV 4:2:0 pixel format before H.266
encoding; objective metrics were computed on the same format.

\label{sec:supp_vp8}
VP8 pre-encoding was used to simulate WebRTC-style capture
conditions~\cite{alvestrand_webrtc_2016}.
The NR-TH benchmarking clips were encoded with VP8 in constant bitrate
(CBR) mode at 2500\,kbps, a representative target for HD video calls.
The VP8-encoded output was decoded back to raw YUV using FFmpeg's
default VP8 decoder and then used as input to the codec benchmarking
pipeline above.

\begin{table*}[!t]
\centering
\caption{Encoder configurations used in the codec benchmarking
experiments.}
\label{tab:supp_encoding}
\begin{tabular}{@{}ll@{}}
\toprule
Encoder & Command \\
\midrule
H.264 &
  \texttt{ffmpeg -init\_hw\_device qsv=hw -filter\_hw\_device hw} \\
(Intel QSV) &
  \texttt{-i \{input\} -c:v h264\_qsv -load\_plugin h264\_hw} \\
&
  \texttt{-scenario 2 -p\_strategy 0 -b\_strategy 0 -threads 1} \\
&
  \texttt{-sc\_threshold 0 -preset fast -bf 0 -look\_ahead 0} \\
&
  \texttt{-g 6000 -i\_qfactor 1.0 -i\_qoffset 0 -b\_qfactor 1.0} \\
&
  \texttt{-b\_qoffset 0 -refs 1 -low\_power 1 -q \{qp\} \{output\}} \\
\midrule
H.265 &
  \texttt{ffmpeg -init\_hw\_device qsv=hw -filter\_hw\_device hw} \\
(Intel QSV) &
  \texttt{-i \{input\} -c:v hevc\_qsv -load\_plugin hevc\_hw} \\
&
  \texttt{-scenario 2 -p\_strategy 0 -b\_strategy 0 -threads 1} \\
&
  \texttt{-sc\_threshold 0 -preset fast -bf 0 -look\_ahead 0} \\
&
  \texttt{-g 6000 -i\_qfactor 1.0 -i\_qoffset 0 -b\_qfactor 1.0} \\
&
  \texttt{-b\_qoffset 0 -refs 1 -low\_power 1 -q \{qp\} \{output\}} \\
\midrule
AV1 &
  \texttt{aomenc --codec=av1 --ivf --i420 --end-usage=q} \\
(libaom 3.13.1) &
  \texttt{--threads=1 --passes=1 --disable-kf --lag-in-frames=0} \\
&
  \texttt{--cpu-used=8 --sb-size=64 --psnr --rt --enable-cdef=1} \\
&
  \texttt{--tune-content=default --cq-level=\{qp\} -o \{output\} \{input\}} \\
\midrule
H.266 &
  \texttt{vvencFFapp -c lowdelay\_fast.cfg -c extra\_vvenc.cfg} \\
(VVenC 1.14.0) &
  \texttt{--InputFile \{input\} -s \{W\}x\{H\} -fr \{fps\}} \\
&
  \texttt{-b \{bitstream\} -o \{output\} --QP \{qp\} --Threads 1} \\
&
  extra\_vvenc.cfg: \texttt{InternalBitDepth:8~~OutputBitDepth:8} \\
&
  \texttt{PicReordering:0~~NumPasses:-1~~LookAhead:-1} \\
\midrule
VP8 &
  \texttt{ffmpeg -i \{input\} -map 0:v:0 -an -c:v libvpx} \\
(libvpx, pre-encode) &
  \texttt{-pix\_fmt yuv420p -deadline realtime -cpu-used 5} \\
&
  \texttt{-lag-in-frames 0 -error-resilient 1} \\
&
  \texttt{-g 60 -keyint\_min 60 -b:v 2500k -minrate 2500k} \\
&
  \texttt{-maxrate 2500k -bufsize 2500k \{output\}} \\
\bottomrule
\end{tabular}
\end{table*}

\end{document}